\documentclass[preprint,12pt]{elsarticle}
\usepackage[margin=1in]{geometry}
\usepackage{xcolor}
\usepackage{tikz}
\usetikzlibrary{shapes,arrows}
\usetikzlibrary{positioning,decorations.pathreplacing}
\usetikzlibrary{backgrounds}
\usepackage{amsmath,bm}
\usepackage{amssymb}
\usetikzlibrary{positioning}
\usepackage{subcaption}
\usepackage{algorithm,algpseudocode}
\usepackage{mathtools}
\usepackage{rotating,caption,booktabs}
\usepackage{float}
\usepackage{hvfloat}
\usepackage{booktabs}
\usepackage{adjustbox}
\usepackage{natbib}
\setcitestyle{authoryear,open={(},close={)}}
\usepackage[normalem]{ulem}
\usepackage{makecell, multirow}
\usepackage{scrextend}
\usepackage{footnote}
\makesavenoteenv{tabular}
\makesavenoteenv{table}

\usepackage{lscape}
\usepackage{lineno}
\newcommand{\red}{\textcolor{black}}
\newcommand{\blue}{\textcolor{black}}

\newcommand{\itemname}{\makebox[2.1cm][l]}
\newcommand{\pmtab}{\multicolumn{1}{@{\hspace*{\tabcolsep}\makebox[0pt]{$\pm$}}c|}}
\newcommand{\pmtabend}{\multicolumn{1}{@{\hspace*{\tabcolsep}\makebox[0pt]{$\pm$}}c}}
\newcommand{\pmtabendbf}{\multicolumn{1}{@{\hspace*{\tabcolsep}\makebox[0pt]{$\bm{\pm}$}}c}}
\newcommand{\pmtabbf}{\multicolumn{1}{@{\hspace*{\tabcolsep}\makebox[0pt]{$\bm{\pm}$}}c|}}

\usepackage{titlesec}
\usepackage{lmodern}
\titleformat*{\subsection}{\fontsize{13}{18}\itshape}

\makeatletter
\def\ps@pprintTitle{%
 \let\@oddhead\@empty
 \let\@evenhead\@empty
 \def\@oddfoot{\centerline{\thepage}}%
 \let\@evenfoot\@oddfoot}
\makeatother

\newcommand\blfootnote[1]{%
 \begingroup
  \renewcommand\thefootnote{}\footnote{#1}%
  \addtocounter{footnote}{-1}%
  \endgroup
}

\begin{document}

\begin{frontmatter}

\title{\red{Enhancing Discrete Choice Models with Representation Learning}}

\author[epfl]{Brian Sifringer}
\author[tue]{Virginie Lurkin}
\author[epfl]{Alexandre Alahi}
\address[epfl]{Visual Intelligence for Transportation Laboratory (VITA), \'Ecole Polytechnique F\'ed\'erale de Lausanne (EPFL), Switzerland.}
\address[tue]{Department of Industrial Engineering \& Innovation Sciences, Eindhoven University of Technology, Eindhoven 5600MB, The Netherlands}

 \begin{abstract}
In discrete choice modeling (DCM), model misspecifications may lead to limited predictability and biased parameter estimates. In this paper, we propose a new approach for estimating choice models in which we divide the systematic part of the utility specification into (i) a knowledge-driven part, and (ii) a data-driven one, which learns a new representation from available explanatory variables. Our formulation increases the predictive power of standard DCM without sacrificing their interpretability. We show the effectiveness of our formulation by augmenting the utility specification of the Multinomial Logit (MNL) and the Nested Logit (NL) models with a new non-linear representation arising from a Neural Network (NN), leading to new choice models referred to as the \textit{Learning Multinomial Logit} (L-MNL) and \textit{Learning Nested Logit} (L-NL) models. Using  multiple publicly available datasets based on revealed and stated preferences, we show that our models outperform the traditional ones, both in terms of predictive performance and accuracy in parameter estimation. All source code of the models are shared to promote open science.
 \end{abstract}
 

\end{frontmatter}


\section{Introduction}
\blfootnote{\small{ \textcopyright 2020. This manuscript version is made available under the CC-BY-NC-ND 4.0 license \url{http://creativecommons.org/licenses/by-nc-nd/4.0/}}}
Discrete Choice Models (DCM) have emerged as a powerful theoretical framework for studying choices made by humans. The goal of these models is to \textit{predict} the choice outcome among a given set of discrete alternatives (\textit{e.g.}, choice of walking as a transportation mode, rather than taking the car or the bus), while \textit{understanding} the behavioral process that led to the specific choice. For many years, the Multinomial Logit model (MNL), based on a simple parametric utility specification, has provided the foundation for the analysis of discrete choice. Despite its oversimplified assumptions regarding the actual decision-making process, MNL is still commonly used in practice since it enables a high level of interpretability. 

Interpretability is critical for researchers and practitioners. For instance, parametric specifications based on linear or logarithmic functions allow for straightforward derivation of the value-of-time (VOT), \textit{i.e.}, the marginal rate of substitution between time and cost, that constitutes a highly relevant measure in a wide range of public transport policy. However, interpretability gain from using an MNL model with such simple parametric specifications often leads to a sacrifice in the predictive power of the model. Indeed, MNL does not allow to incorporate individual heterogeneity explicitly in the empirical investigation, and the assumed simple parametric model cannot adequately capture the full underlying structure of the data. Several prior studies, within the discrete choice literature, have shown that incorrectly assuming a parametric utility specification can cause severe bias in parameter estimates and post-estimation indicators (\textit{e.g.}, \cite{torres2011,van2014}). 

More advanced utility specifications, embedded into more complex models, would allow for a better fit of the data and, ultimately, a better prediction. In this regard, utility specifications based on Box-Cox transformation, piecewise-linear, exponential, or non-parametric functions have been proposed in the literature (\textit{e.g.}, \citet{schindler2007,kneib2007,kim2016}), as well as advanced DCM models, such as the Mixed Logit Model (MLM) or the Latent Class Model (LCM)  (\textit{e.g.}, \citet{shen2009,xiong2013,vij2013,kim2016}). However, the main limitation of these choice models is that the standard procedure for estimating the parameters' values is to assume that the model specification is known \textit{a priori}, while in practice determining the utility specification for a particular application remains a difficult task.

Within the machine learning community, Neural Networks (NN) masters the field of representation learning. They have become the go-to solutions, as their superior prediction performance has been repeatedly demonstrated, including in transportation applications (\textit{e.g.},  \citet{omrani2015,hagenauer2017,pekel2017,nam2017,pirra2018, chang2019travel,zhou2019bike}). Unlike discrete choice models, NN requires essentially no \textit{a priori} beliefs about the nature of the true underlying relationships among variables. However, gaining in prediction accuracy often comes at the cost of losing interpretability, explaining why the discrete choice community often considers them as a black-box. 

In this work, we propose a new approach to discrete choice modeling by integrating representation learning in the formulation.  We propose to divide the systematic part of the utility specification into (i) a knowledge-driven (interpretable) part and (ii) a data-driven (representation learning) part, which aims at automatically discovering a good utility specification from available data. Our formulation partially replaces manual utility specification and allows for a better prediction performance without sacrificing interpretability. We demonstrate the effectiveness of our framework by augmenting the utility specification of the MNL and Nested logit models with a new non-linear representation, arising from a neural network. It leads to new choice models, referred to as the \textit{Learning Multinomial Logit} (L-MNL) and \textit{Learning Nested Logit} (L-NL) models, respectively. The source code is made publicly available to ease reproducibility and promote open science\footnote{{https://github.com/BSifringer/EnhancedDCM}}. 

The remainder of the paper is organized as follows: Section \ref{sec:related} is a non-exhaustive overview of the recent attempts to conduct multi-disciplinary research, combining machine learning (ML) and discrete choice modeling. Section \ref{sec:NN and DCM} gives a brief background and presents details on how we implement MNL with modern neural network libraries. Section \ref{sec:L-MNL} describes our proposed L-MNL and L-NL models. Section \ref{sec:experiments} demonstrates evidence that our learning choice models outperform both existing hybrid neural network and MNL models based on  simple parametric utility specifications. \color{black} Our results on real and synthetic data show that the research community can benefit from a better integration of representation learning techniques and discrete choice theory. In that sense, our research also contributes to bridging the gap between knowledge-driven and data-driven methods. 

\section{Related work}\label{sec:related}

Discrete choice and machine learning researchers have typically different perspectives and research priorities. The related work in each field is too broad to be covered here exhaustively, and we, therefore, limit ourselves to the most relevant works at the intersection of the two communities. 

More than 20 years ago, researchers were already interested in applying data-driven methods such as neural networks in different transportation applications (\textit{e.g.}, \citet{faghri1992,dougherty1995,shmueli1996}). Over the years, different types of NN, as well as decision trees, have been compared to MNL models (\citet{agrawal1996,lee2018comparison, zhao2020prediction}), Nested Logit (NL) models (\citet{mohammadian2002nested,hensher2000comparison}) and more generally to random utility models (\citet{sayed2000comparison,cantarella2005multilayer,paredes2017machine}) and statistical methods (\citet{west1997comparative,karlaftis2011statistical,iranitalab2017comparison,golshani2018modeling,brathwaite2017machine}). However, the literature has mainly focused on comparing the models in terms of prediction power, without discussing the issue of interpretability. 

The machine learning community has also generated a tremendous amount of research aiming at predicting, with high accuracy, a variety of choices. Among the most recent works, \cite{pekel2017} or \cite{jin2020systematic} present comprehensive reviews of publications related to the application of NN or ML to predict travelers' choice in public transportation. \cite{hagenauer2017} compare seven machine learning methods to classify travel mode choice. The methods investigated are MNL and NN, but also Naive Bayes (NB) (\cite{rish2001}), Gradient Boosting Machines (GBM) (\cite{friedman2001}), Bagging (BAG) (\cite{breiman1996}), Random Forests (RF) (\cite{breiman2001}), and Support Vector Machine (SVM) (\cite{cortes1995}). \cite{pirra2018} also propose to use SVM for travel mode choice while \cite{lheritier2018} rely on RF and GBM to predict airline itinerary choice. Not surprisingly, the results exhibited in these papers show outstanding results in the ability of these models to fit the data but do not tackle the issue of their interpretability.

Some recent publications have gone beyond the comparison of the two fields and proposed innovative behavioral studies based on data-driven methods. Two notable examples are the work of \cite{wong2018}, who use a restricted Boltzmann Machine (BM) (\cite{ackley1985}) to represent latent behavior attributes, and the study of \cite{Sander2019}, who develop a novel NN based approach to investigate decision rule heterogeneity amongst travelers. {Moreover, \cite{wang2018using} have investigated the numerical extraction of behavioral information such as willingness to pay in a Dense Neural Network (DNN) using elasticity while \cite{wang2018framing} have investigated the theoretical transition from DCM to DNN and subsequent trade between predictability and interpretability.} {Recently, the main authors have also developed a new structuring of neural networks, based on alternative specific utility architecture (\cite{wang2020deep}). Finally, \cite{wong2020bi} have made use of a NN generative model to capture the joint distribution of multiple discrete-continous travel data.} However, these papers do not have the objective of finding a utility specification that allows high predictability while maintaining straightforward interpretability. 

The closest studies relative to our work are in the brand choice literature. \citet{bentz2000neural} introduce a hybrid model, afterward called the NN-MNL model. Their model involves a two-stage approach that starts with the estimation of a NN model that aims at discovering non-linearity effects in the utility function. Then, if the NN identifies some non-linearities, the specification of the MNL model is modified to include new variables, specially created to account for the discovered non-linear effects. On their dataset, the re-specified MNL model, which includes the discovered non-linear effects, slightly outperforms the basis MNL model. Their work, similar to ours in spirit, aims at achieving better predictive power together with a greater understanding of the influencing factors. However, their approach is sequential, and their neural network is only used beforehand as a diagnostic and specification tool. Again, in the context of brand choice, \citet{hruschka2002flexible,hruschka2004empirical} and \citet{hruschka2007using} investigate further the potential of neural network based choice models. In \cite{hruschka2002flexible}, the NN-MNL model is compared against homogeneous and heterogeneous versions of linear utility MNL models. Meanwhile, in \cite{hruschka2004empirical}, the comparison is made against two non-linear models, the generalized additive (GAM-MNL) model of \cite{abe1999generalized} and a flexible functional form based on Taylor series approximations. The authors show that, by being capable of finding non-linear relationships from available data, the neural network based choice models with non-linear specifications outperform the other models in terms of prediction performance. However, the authors are unable to draw any conclusion regarding the significance of the factors with this method. Also, the post estimation analysis is limited to likelihood and choice elasticities. Finally, \cite{hruschka2007using} proposes a Multinomial Probit (MNP) model with a neural network extension to model brand choice. Their model combines the heterogeneity across households with a multilayer perceptron. Linear and non-linear deterministic utility specifications are considered by specifying the same variables in linear and non-linear terms, hence loosing interpretability. The probit assumption also prevents the closed form of the choice probabilities and the model can only be estimated using a Markov Chain Monte Carlo simulation technique.  

Our work significantly departs from these previous studies by proposing a flexible and general framework for the specification of the deterministic utility in any DCM model. More specifically, our approach combines, in a single joint optimization, the estimation of a standard DCM model, based on a carefully thought-out utility specification, with a representation learning model that aims at increasing the overall predictive performance. To our knowledge, this is the first formulation that benefits from the predictive power of representation learning techniques while keeping some key parameters interpretable, which allows us to derive insightful post-estimation indicators. In Section \ref{sec:experiments}, we show the effectiveness of our learning choice method by comparing its performance with benchmarking models, including the NN-MNL model of \citet{hruschka2002flexible,hruschka2004empirical}.

\section{Background and Notations}\label{sec:NN and DCM}

\subsection{Multinomial Logit} \label{sec:MNL}

Discrete choice modeling complies with the Random Utility Maximization (RUM) theory (\cite{mcfadden1974}), which postulates that an individual is a rational decision-maker, who aims at maximizing the utility relative to their choice. Utility is a latent construct assumed to be partitioned into two components: a systematic (or deterministic) utility, $V_{in}$, and a random component, $\varepsilon_{in}$, that captures the uncertainty coming from the impossibility for the modeler to fully capture the choice context. Formally, the utility that individual $n$ associates with alternative $i$ from their choice set $\mathcal{C}_n$ is given as\footnote{{Notation convention can be found in \ref{sec:notation}.}}:
\begin{equation} \label{eq:standardutility}
 U_{in} =  V_{in} + \varepsilon_{in},
\end{equation}

For convenience, we consider the systematic part of the utility to be linear-in-parameter, as it is generally assumed: 
\begin{equation} \label{eq:deterministic}
  V_{in}
 = \sum_{d} \beta_d \cdot x_{din}, 
 \end{equation}
\noindent where $\bm{\beta}$ are the preference parameters (or estimators) associated with the explanatory variables (or the input features) $\bm{x}\in\mathcal{X}$ that describe the observed attributes of the choice alternative (\textit{e.g.}, the price or travel time associated with the mode), and the individual's socio-demographic characteristics (\textit{e.g.},  the individual's level of income or age). 
  
Under the standard MNL assumption that error terms are independently and identically distributed (i.i.d.) and follow an Extreme Value Type I distribution with location parameter zero and the scale parameter 1  (\textit{i.e.} $\varepsilon_{in} \overset{\text{i.i.d.}}{\sim} EV(0, 1)$), the probability for individual $n$ to select choice alternative $i$ is given by
\begin{equation} \label{eq:MLproba}
P_n(i) = \frac{e^{V_{in}}}{\sum_{j\in \mathcal{C}_n} e^{V_{jn}}}. 
\end{equation}
The preference parameters $\bm{\beta}$ are typically estimated by maximizing the log-likelihood function given by:
 \begin{equation} \label{eq:logloss}
 \mathcal{L} =  \sum\limits_{n=1}^{N}\sum\limits_{i\in \mathcal{C}_n} y_{in} \log\left[P_n(i)\right],
 \end{equation}
\noindent where $y_{in}$ is the observed choice variable (or true label) and is equal to $1$ if the choice of individual $n$ is the alternative $i$, and to $0$ otherwise.

\subsection{\red{Implementing MNL as a Neural Network }}

A neural network consists of a function mapping the input space $x$ to an output of interest $U$ through several intermediate representations commonly referred to as hidden layers $\bm{h}^{(j)}$: 
\begin{alignat}{3}
 && \bm{U} & =  \bm{h}^{(L)}(\bm{q}^{(L-1)}),\\
\hspace{-1cm} with \hspace{1cm} &&\bm{q}^{(j)} & =  \bm{h}^{(j)}(\bm{q}^{(j-1)}), \hspace{1cm} \forall j=1,...,L, \hspace{-3cm}
\end{alignat}
\noindent where $\bm{q}^{(0)}=\bm{x}$ and $L$ is the last representation layer.

We make use of a Convolutional Neural Network (CNN) to retrieve the MNL formulation\footnote{Previous studies, such as \cite{bentz2000neural}, have successfully written MNL with other Neural Network architectures. We have used CNN for its weight-sharing architecture.}. A CNN has weights in the shape of a filter that connects a layer $\bm{h}^{(j)}$ to the next by applying a convolution\footnote{{ This method is often seen in image processing where a filter with a fix number of weights is applied to an equal amount of inputs by multiplying the terms together and then summing them over to obtain a single new value. A new image is obtained by sliding the filter over all inputs with a set step-size named stride.  In our case, the operation is done without padding, \textit{i.e.,} without applying the filter out of the boundaries, effectively reducing the size from one layer (filtering) to the next. In image processing, the correlation formula is the same as a convolution up to a flip in the order of its entries. Without loss of generality, we write the correlation formula in Equation \ref{eq:cnn} for readability.}}. Therefore, the value of a neuron $i$ in the next layer $(j+1)$ can be written as:
\begin{equation}\label{eq:cnn}
h^{(j+1)}_i = g(\sum\limits_{k=0}^{d} h^{(j)}_{(s\cdot i+ k)}\beta_{k}^{(j)} + \alpha_i^{(j)}),
\end{equation}

\noindent where $\{\beta_1,...\beta_d \} = \bm{\beta} $ is the filter of size $(1\times d)$, $s$ the stride of the convolution, $\alpha_i$ a bias term and $g(\cdot)$ an activation function.

The MNL formulation is retrieved by using a single layer ($L=1$), setting the activation function to identity ($g(x)=x$) and the stride $s$ to $d$. Doing so, we get the utility functions $\textbf{V}_n = \{V_{1n},...V_{In}\}$ as defined in Equation (\ref{eq:deterministic}).

Then, the probabilities can be obtained by using a softmax activation layer (\cite{bishop1995neural}) defined as:
\begin{equation}\label{eq:softmax}
 (\bm{\sigma}(\textbf{V}_n))_i =  \frac{e^{V_{in}}}{\sum_{j\in \mathcal{C}_n} e^{V_{jn}}}, 
 \end{equation}
which can be identified as Equation (\ref{eq:MLproba}) for all probabilities. The output of the network goes through a loss function, in our case categorical cross-entropy (CE)(\cite{shannon1948mathematical}):
\begin{equation}\label{CE}
 H_n(\bm{\sigma},\textbf{y}_n) = - \sum\limits_{i\in\mathcal{C}_n} y_{in} \log\left[\sigma_i(\textbf{V}_n)\right].
 \end{equation}
Minimizing (\ref{CE}) is equivalent to maximizing Equation (\ref{eq:logloss}) when summed over all individuals $n$.
 
 
 \tikzstyle{naveqs} = [sensor, text width=7em, opacity=.2, fill=blue!20, 
    minimum height=30em, rounded corners]
    \tikzstyle{sensor}=[draw, fill=black!20, text width=3em, 
    text centered, minimum height=2.5em]

\usetikzlibrary{backgrounds}
\def\layersep{3cm}
\begin{figure}[t]
\begin{center}
\begin{tikzpicture}[scale=0.75,shorten >=1pt,->,draw=black!50, node distance=\layersep]
    \tikzstyle{every pin edge}=[<-,shorten <=1pt]
    \tikzstyle{neuron}=[circle,fill=black!25,minimum size=17pt,inner sep=0pt]
    \tikzstyle{input neuron 1}=[neuron, fill=red!25];
    \tikzstyle{input neuron 2}=[neuron, fill=blue!25];
    \tikzstyle{input neuron 3}=[neuron, fill=green!25];
    \tikzstyle{output neuron}=[neuron, fill=black!100];
    \tikzstyle{hidden neuron}=[neuron, fill=black!50];
    \tikzstyle{void neuron}=[neuron, fill=white!50, draw=black!50];
    \tikzstyle{annot} = [text width=4em, text centered]

        \node[input neuron 1, pin=left: ] (I-1) at (0,-1) {};
        \node[input neuron 1, pin=left: ] (I-2) at (0,-2) {};
        \node[input neuron 2, pin=left: ] (I-3) at (0,-3) {};
        \node[input neuron 2, pin=left: ] (I-4) at (0,-4) {};
        \node[input neuron 3, pin=left: ] (I-5) at (0,-5) {}; 
        \node[input neuron 3, pin=left: ] (I-6) at (0,-6) {};        

\node (car) at (-3.5,-1.5) {Car};
	\begin{scope} [node distance = .5em]
		\node [right=of car, yshift=1em] (co) {cost};
		\node [right=of car, yshift=-0.8em] (ti) {time};
	\end{scope}
	\draw[decorate,decoration={brace, mirror}] (co.north west) -- (ti.south west);        
\node (train) at (-3.7,-3.5) {Train};        
	\begin{scope} [node distance = .5em]
		\node [right=of train, yshift=1em] (co) {cost};
		\node [right=of train, yshift=-0.8em] (ti) {time};
	\end{scope}
	\draw[decorate,decoration={brace, mirror}] (co.north west) -- (ti.south west);

\node (sm) at (-3.5,-5.5) {SM};        
	\begin{scope} [node distance = .5em]
		\node [right=of sm, yshift=1em] (co) {cost};
		\node [right=of sm, yshift=-0.8em] (ti) {time};
	\end{scope}
	\draw[decorate,decoration={brace, mirror}] (co.north west) -- (ti.south west);

        \path[yshift=0.5cm]
            node[hidden neuron, pin={[pin edge={draw=none}, pin distance = 0.15cm, yshift=0.2cm, xshift = 0.2cm]left:$\beta_{c}$}] (H-1) at (\layersep,-1 cm) {};
        \path[yshift=0.5cm]
            node[hidden neuron, pin={[pin edge={draw=none}, pin distance = 0.15cm, yshift=0.2cm, xshift = 0.2cm]left:$\beta_{t}$}] (H-2) at (\layersep,-2 cm) {};

	\foreach \name / \y in {3,4}
        \path[yshift=0cm]
            node[void neuron] (H-\name) at (\layersep,-\y cm) {};	\foreach \name / \y in {5,6}
        \path[yshift=-0.5cm]
            node[void neuron] (H-\name) at (\layersep,-\y cm) {};
            
    \draw (\layersep-0.5cm, 0 cm) rectangle (\layersep+0.5cm, -2cm);         
    \draw [dashed] (\layersep-0.5cm, -2.5 cm) rectangle(\layersep+0.5cm, -4.5cm);
    \draw [dashed] (\layersep-0.5cm, -5 cm) rectangle (\layersep+0.5cm, -7cm);         

    \node[output neuron, fill=red!40, minimum size=27pt] (O)  at (2*\layersep,-2 cm) {$V_{car}$};
    \node[output neuron, fill=blue!40] (p)  at (2*\layersep,-3.5 cm) {$V_{train}$};
    \node[output neuron, fill=green!40, minimum size=27pt] (q)  at (2*\layersep,-5 cm) {$V_{SM}$};

    \foreach \source in {1,2}
         \path (I-\source) edge (H-\source);
    \foreach \source in {3,...,6}
         \path (I-\source) edge [dashed](H-\source);
    \path (H-2) edge (H-3);
    \path (H-4) edge (H-5);
    
    \foreach \source in {1,2}
        \path (H-\source) edge (O);
    \foreach \source in {3,4}
        \path (H-\source) edge [dashed] (p);
    \foreach \source in {5,6}
        \path (H-\source) edge [dashed] (q);


    \node (loss) at (4*\layersep-1cm, -3.5cm) {CE Loss};
	

 	\draw[fill=white] (3*\layersep-1.5cm, -1 cm) rectangle node{{\rotatebox{90}{Softmax}}}(3*\layersep-0.5cm, -6cm);
 	\begin{scope}[on background layer]
 		\path (O) edge[draw=black!50] (loss);
    		\path (p) edge[draw=black!50] (loss);
   		\path (q) edge[draw=black!50] (loss);
 	\end{scope}

    \node[annot,above of=H-1, node distance=1cm, xshift=-0.2cm] (hl) {Filter};
    \node[annot,left of=hl, node distance=\layersep-1cm] {Input layer $\bm{x}$};
    \node[annot,right of=hl, node distance=\layersep-1.1cm] (ol) {Hidden \hspace*{-0.1cm}layer $\bm{h}^{(1)}$\hspace*{-0.1cm}};
    \node[annot,right of=ol, node distance=\layersep-1.1cm] (af) {Activation function};  
    \node[annot,right of=af, node distance=\layersep-1.1cm] {Output layer};  
\end{tikzpicture}
\caption{By aligning inputs by class and convolving with a filter of equivalent shape and stride, we can retrieve linear utility specifications with a single CNN layer.
By ending the network with a softmax activation layer and a cross-entropy (CE) loss, we retrieve the same formulation as for the MNL model. \label{fig:CNN}}
\end{center}
\end{figure}
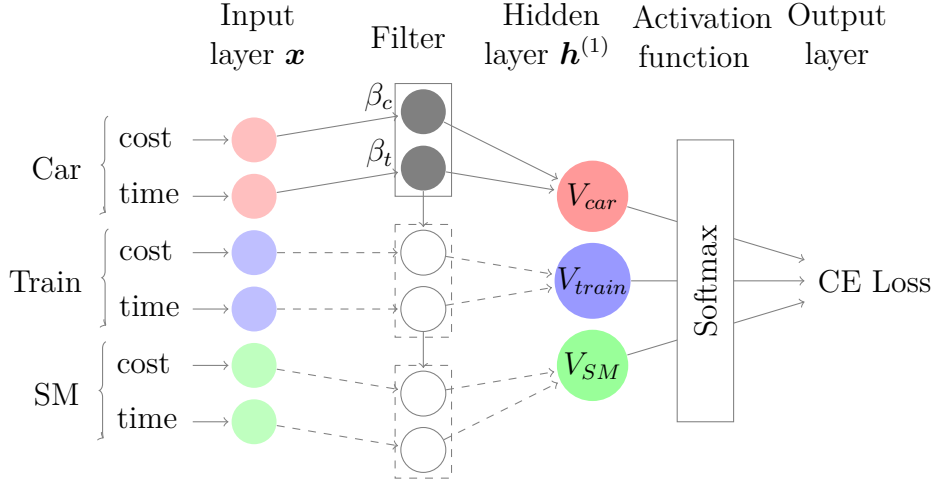
 
 An illustrative example is given in Figure \ref{fig:CNN} with :
 \begin{align}
 U_{in} &= \beta_{c }\cdot x_{1i} +  \beta_{t} \cdot x_{2i} + \varepsilon_{in},\hspace{0.25cm}  \forall i\in \mathcal{C},
 \end{align}
\noindent where $x_{1} = cost$ stands for travel cost, and $x_{2} = time$ for travel time. The choice set $\mathcal{{C}}$ is assumed to be the same for all individuals and contains the following mode alternatives: $\mathcal{{C}} = \{\text{Car, Train, {Swissmetro (SM)}}\}$.  The {filter} is made of $\bm{\beta} =(\beta_t,\beta_c$), and a stride of $d=2$ allows to recover each utility specification in the next layer from $cost$ and $time$ of every alternative.


\section{Representation Learning in Discrete Choice Modeling}\label{sec:L-MNL}
\subsection{General Formulation} \label{sec:formulation}
As previously mentioned, the standard statistical procedure for estimating the parameter values of choice models is to assume that the true utility specification is known \textit{a priori}. Typically, the most common representation is inspired from linear regression and involves the observed attributes of the choice alternative and the individual's socio-demographic characteristics. However, the influence of these explanatory variables on the utility of a choice alternative is unlikely to be known and to be precisely linear. The danger of an incorrect utility specification remains, therefore, highly present. 

In this work, we propose a more flexible and data-driven approach that consists in expressing the systematic part of the utility function into two sub-parts, as follows:
 \begin{equation} \label{eq:systematicutility}
 V_{in} =  f_{i}(\mathcal{X}_n;\bm{\beta}) + r_{i}(\mathcal{Q}_n, \bm{w}),
\end{equation} 
\noindent where 
\begin{itemize}
\item[-] $\bm{f}(\mathcal{X}_n;\bm{\beta})$
 is the knowledge-driven part, \color{black} assumed interpretable. The function $f$ is defined such that its unknown model parameters $\bm{\beta}$ are an interpretable combination of the explanatory variables (or input features) $\mathcal{X}_n$.  
\item[-]$ \bm{r}(\mathcal{Q}_n, \bm{w})$   is the data-driven part,  \color{black} a representation that is learned from a set of explanatory variables (or input features) $\mathcal{Q}_n$, for which no \textit{a priori} relationship is assumed.
\end{itemize}

Conditions between $\mathcal{X}$ and $\mathcal{Q}$ to keep $\bm{f}$ interpretable are described in section $\ref{sec:modeling}$.
 
Replacing the systematic utility component in Equation (\ref{eq:standardutility}) by its new expression given by Equation (\ref{eq:systematicutility}), we obtain the following utility expression: 
 \begin{equation} \label{eq:ourutility}
  \bm{U}_{n} =  \bm{f}(\mathcal{X}_n;\bm{\beta}) + \bm{r}(\mathcal{Q}_n;\bm{w}) + \bm{\varepsilon}_{n}.
 \end{equation}
\red{
We may interpret this equation as a data-driven term finding the residual of a hand-modeled function. \cite{he2016deep} have shown the impact of residual network (ResNet) in the machine learning community{, and its particular architecture has been widely used for increased interpretability by finding the residual of knowledge and equation based modeling (\cite{liao2018dest,wang2018enresnet,li2019deep}), or choice modeling (\cite{wong2019reslogit})}. Our work differs with the use of two input sets $\mathcal{X}$ and $\mathcal{Q}$ necessary for keeping discrete choice interpretability {as we define it in section \ref{sec:modeling} and further explain in section \ref{sec:experiments}}.
}

\red{Finally, the above specification may be applied to any discrete choice modeling kernel. However, the following applications will limit themselves to models with Gumbel distributed random terms.}

\subsection{{Modeling}}
\label{sec:modeling}

When modeling a utility specification with the formulation in Equation (\ref{eq:ourutility}), one must understand the attributes of a parameter related to either $\mathcal{X}$ in $\bm{f}$ or $\mathcal{Q}$ in $\bm{r}$. The first assumes to have all the benefits of expert modeling in discrete choice, which includes the availability of the $\bm{\beta}$ parameters' Hessian and thus their standard deviation approximation. The second produces a new representation of its associated input, which can span from discrete to continuous inputs such as signals and images, while the hessian is generally not available due to computational complexity. 
Therefore, when one studies the impact, or elasticity, of a feature $t_{in}$ on alternative $i$, this readily translates as: 
\begin{equation} \label{eq:elasticity}
\frac{\partial U_{in}}{\partial t_{in}} = \frac{\partial f_{in}}{\partial t_{in}} + \frac{\partial r_{in}}{\partial t_{in}}
\end{equation}

\cite{bentz2000neural} and \cite{wang2018using} have shown that behavioral insight can be captured and studied through the elasticities of a neural network. However, to ensure the same straightforward interpretability of behavioral choice models for a parameter $t$, we explicitly stipulate that the standard deviation must be available, and the elasticity must not depend on the representation term. In other words, we impose that: 
\begin{equation} \label{eq:criticalassumption}
\frac{\partial r_{in}}{\partial t_{in}} = 0
\end{equation}

Equation (\ref{eq:criticalassumption}) constitutes a critical assumption which ensures that insightful post-estimation indicators can be retrieved. This definition, as well as the general formulation is what sets us apart from old and new literature alike. Indeed, to only cite a few, \cite{hruschka2007using} or \cite{wong2018} are specific cases of our formulation when both sets are the same, \textit{i.e.}, $\mathcal{X}=\mathcal{Q}$, while \cite{bentz2000neural}, \cite{otsuka2016deep} and \cite{wang2018using} are cases where $\mathcal{X}=\emptyset$. All of which do not follow our goal of fixing a biased expert modeling component by leveraging the strength of data-driven methods in an added and non-overlapping representation term.

Moreover, the global theoretical effect of this new term may be derived. When one considers Equation (\ref{eq:ourutility}) without the new term, then this value would be found in the random component such that:
\begin{equation} \label{eq:error_term}
    \bar{\varepsilon}_{in} = {r}_i(\mathcal{Q}_n;\bm{w}) + \varepsilon_{in}.
\end{equation}

\noindent To have unbiased parameter estimates in $\bm{f}(\mathcal{X}_n;\bm{\beta})$, one must avoid correlation between the specification and the random terms known as endogeneity, avoid correlation of the random terms between each alternative and overall avoid utility misspecification. Many methods have been developed to specifically fix for endogeneity, such as \cite{guevara2015critical}, while others account for correlation between alternatives (see section \ref{sec:nests}). Our model aims at correcting for underfit due to misspecification and omitted variable bias thanks to data-driven methods. In this sense, the new error term $\varepsilon_{in}$ is more likely to fit the aforementioned criteria  than $\bar{\varepsilon}_{in}$. The new representation term increases the probability of having unbiased parameters while increasing the prediction rate as it may easily find undiscovered specifications during the expert modeling process.  Moreover, our general framework does not exclude the coupling with other methods that aim at fixing the random term, as we will observe in the following sections.

\subsection{L-MNL Model Formulation} \label{sec:L-MNLformulation}

{In the specific case where we choose to add a representation term to a Multinomial Logit for Equation (\ref{eq:ourutility})}, the standard MNL assumptions regarding the distribution of the error term $\varepsilon_{in}$ (described in Section \ref{sec:MNL}) have to be made.

The likelihood of selecting the choice alternative $i$ for individual $n$, given the values of the model parameters ($\bm{\beta}$ and $\bm{w}$) and the influencing factors ($\mathcal{X}_n$ and $\mathcal{Q}_n$), is therefore naturally expressed as: 
\begin{equation} \label{probabilities}
P_n(i) = \frac{e^{f_i(\mathcal{X}_n;\bm{\beta}) + r_i(\mathcal{Q}_n;\bm{w})}}{\sum_{j\in \mathcal{C}_n} e^{f_j(\mathcal{X}_n;\bm{\beta}) + r_j(\mathcal{Q}_n;\bm{w})}}. 
\end{equation}

For our L-MNL model, we choose to use a Dense Neural Network (DNN) as the learning method. More specifically, its representation term $r_{in}$ is the resulting function of a DNN with $L$ layers of $H$ neurons and a single output per utility function: 
\begin{equation}\label{eq:NN_component}
r_{in} = \sum\limits_{k=1}^H w^{(L)}_{ik} g(\bm{q}_n^{(L-1)}\bm{w}^{(L-1)}_k + \alpha_k^{(L-1)}) + \alpha_i^{(L)},
\end{equation}

\noindent where $g(\cdot)$ is the rectifier linear units (ReLU) activation function and $\bm{q}_n^{(j)}$ is recurrently defined by:
\begin{equation}
\left[\bm{q}_n^{(j+1)}\right]_i = \sum\limits_{k=1}^H w^{(j+1)}_{ik} g(\bm{q}_n^{(j)}\bm{w}^{(j)}_k + \alpha_k^{(j)}) + \alpha_i^{(j+1)},
	\end{equation}

\noindent with $\bm{q}_n^{(0)}$ being the vector of input features $\mathcal{Q}_n$.

The architecture of our L-MNL hybrid model is shown in Figure \ref{fig:hyb}, where we see both the linear component, written as a CNN (see Section \ref{sec:NN and DCM}), and the learned representation term, obtained through a DNN. {Our hybrid approach is both general and flexible. Other functions of $\bm{f}$ and other architectures for $\bm{r}$ may be chosen. For more complex DCM frameworks, new assumptions must be made on the error terms, probabilities, and likelihood function. For instance, an extension of our work to the nested logit model is shown in the next section. 
}
\def\layersep{3cm}
\begin{center}
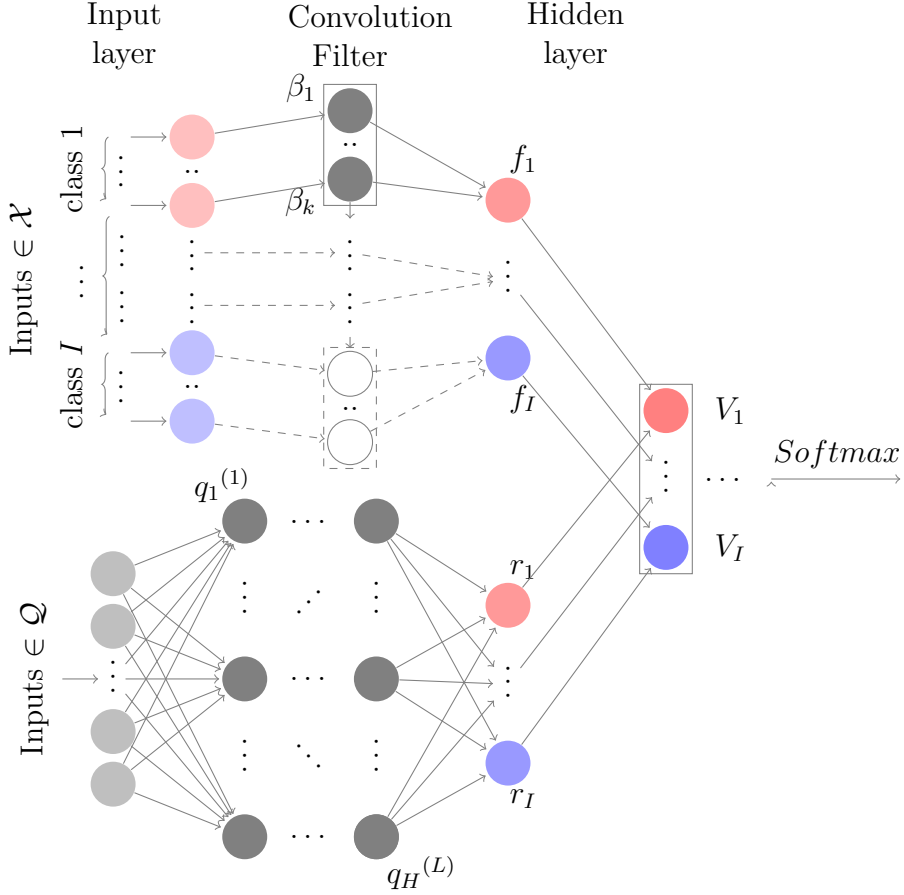
\begin{figure}[h]
\centering
\begin{tikzpicture}[scale=0.7,shorten >=1pt,->,draw=black!50, node distance=\layersep]
    \tikzstyle{every pin edge}=[<-,shorten <=1pt]
    \tikzstyle{neuron}=[circle,fill=black!25,minimum size=17pt,inner sep=0pt]
    \tikzstyle{input neuron 1}=[neuron, fill=red!25];
    \tikzstyle{input neuron 2}=[neuron, fill=blue!25];
    \tikzstyle{input neuron 3}=[neuron, fill=blue!25];
    \tikzstyle{output neuron}=[neuron, fill=black!100];
    \tikzstyle{hidden neuron}=[neuron, fill=black!50];
    \tikzstyle{void neuron}=[neuron, fill=white!50, draw=black!50];
    \tikzstyle{annot} = [text width=4em, text centered]

        \node[input neuron 1, pin=left: ] (I-1) at (0,-0.8) {};
        \node[input neuron 1, pin=left: ] (I-2) at (0,-2.1) {};
        \node[] (I-3) at (0,-3) {\rotatebox{90}{\ldots}};
        \node[] (I-4) at (0,-4) {\rotatebox{90}{\ldots}};
       
        \node[input neuron 3, pin=left: ] (I-5) at (0,-4.9) {}; 
        \node[input neuron 3, pin=left: ] (I-6) at (0,-6.2) {};

\node (car) at (-2.3,-1.4) {\rotatebox{90}{class 1}};
	\begin{scope} [node distance = .5em]
        \node [right=of car] (class1) {\rotatebox{90}{\ldots}};
	\end{scope}
	\draw[decorate,decoration={brace, mirror}] (class1.north west) -- (class1.south west);        
\node (train) at (-2.1,-3.5) {\rotatebox{90}{\ldots}};        
	\begin{scope} [node distance = .5em]
		\node [right=of train, yshift=1em] (co) {\rotatebox{90}{\ldots}};
		\node [right=of train, yshift=-0.8em] (ti) {\rotatebox{90}{\ldots}};
	\end{scope}
	\draw[decorate,decoration={brace, mirror}] (co.north west) -- (ti.south west);

\node (sm) at (-2.3,-5.5) {\rotatebox{90}{class $I$}};        
	\begin{scope} [node distance = .5em]
        \node [right=of sm] (class1) {\rotatebox{90}{\ldots}};
	\end{scope}
	\draw[decorate,decoration={brace, mirror}] (class1.north west) -- (class1.south west);        
\node (inputs) at (-3.2,-3.5) {\rotatebox{90}{Inputs $\in\mathcal{X}$}};

        \path[yshift=0.5cm]
            node[hidden neuron, pin={[pin edge={draw=none}, pin distance = 0.2cm, yshift=0.3cm, xshift = 0.2cm]left:$\beta_1$}] (H-1) at (\layersep,-0.8 cm) {};
        \path[yshift=0.5cm]
            node[hidden neuron, pin={[pin edge={draw=none}, pin distance = 0.2cm, yshift=-0.3cm, xshift = 0.2cm]left:$\beta_k$}] (H-2) at (\layersep,-2.1 cm) {};

	\foreach \name / \y in {3,4}
        \path[yshift=0cm]
            node[] (H-\name) at (\layersep,-\y cm) {\rotatebox{90}{\ldots}};

	\path[yshift=-0.5cm]
            node[void neuron] (H-5) at (\layersep,-4.8 cm) {};
     \path[yshift=-0.5cm]
            node[void neuron] (H-6) at (\layersep,-6.1 cm) {};
            
    \draw (\layersep-0.5cm, 0.2cm) rectangle (\layersep+0.5cm, -2.1cm);         
    \draw [dashed] (\layersep-0.5cm, -4.8 cm) rectangle (\layersep+0.5cm, -7.1cm);

	\path (I-1) -- node[]{..} (I-2);
	\path (I-5) -- node[]{..} (I-6);

	\path (H-1) -- node[]{..} (H-2);
	\path (H-5) -- node[]{..} (H-6);

    \node[output neuron, fill=red!40,  pin={[pin edge={draw=none}, pin distance = 0.005cm, xshift=+0.2cm, yshift=-0.1cm]above:$f_1$}] (o)  at (2*\layersep,-2 cm) {};
    \node[] (p)  at (2*\layersep,-3.5 cm) {\rotatebox{-90}{\ldots}};
    \node[output neuron, fill=blue!40,  pin={[pin edge={white, -, draw=none}, pin distance = 0.005cm, xshift=+0.2cm, yshift=+0.1cm]below:$f_I$}] (q)  at (2*\layersep,-5 cm) {};

    \foreach \source in {1,2}
         \path (I-\source) edge (H-\source);
    \foreach \source in {3,...,6}
         \path (I-\source) edge [dashed](H-\source);
    \path (H-2) edge [] (H-3);
    \path (H-4) edge [] (H-5);
    
    \foreach \source in {1,2}
        \path (H-\source) edge (o);
    \foreach \source in {3,4}
        \path (H-\source) edge [dashed] (p);
    \foreach \source in {5,6}
        \path (H-\source) edge [dashed] (q);

    \node[annot,above of=H-1, node distance=1cm] (hl) {Convolution Filter};
    \node[annot,left of=hl] {Input layer};
    \node[annot,right of=hl] {Hidden layer};
    \node[output neuron, fill=red!50, pin={[pin edge={white}, pin distance=0.2cm]right:$V_{1}$}] (V1) at (3*\layersep, -6cm) {};

    \node[pin={[pin edge={white}, pin distance=0.2cm]right:$\ldots$}] (V2) at (3*\layersep, -7.3cm) {\rotatebox{-90}{\ldots}};

    \node[output neuron, fill=blue!50, pin={[pin edge={white}, pin distance=0.2cm]right:$V_{I}$}] (V3) at (3*\layersep, -8.6cm) {};

\draw (3*\layersep-0.5cm, -5.5cm) rectangle (3*\layersep+0.5cm, -9.1cm);

\path (o) edge (V1); 
\path (p) edge (V2);
\path (q) edge (V3);

\draw (4*\layersep-1cm, -7.3cm) edge node[auto] {\(Softmax\)} (4*\layersep+1.5cm, -7.3cm);
\begin{scope}[shift={(0,-7.7)}]
    \tikzstyle{every pin edge}=[<-,shorten <=1pt]
    \tikzstyle{neuron}=[circle,fill=black!25,minimum size=17pt,inner sep=0pt]
    \tikzstyle{input neuron}=[neuron, fill=black!25];
    \tikzstyle{output neuron}=[neuron, fill=black!100];
    \tikzstyle{hidden neuron}=[neuron, fill=black!50];
    \tikzstyle{annot} = [text width=4em, text centered]

    	\path[yshift = -0.4cm, xshift=-1.5cm] node[ pin={left: {\rotatebox{90}{Inputs $\in \mathcal{Q}$}}}] (I-3) at (0,-3) {\rotatebox{-90}{\ldots}};
    \foreach \name / \y in {1,2,4,5}
    	\path[yshift = -0.4cm,xshift=-1.5cm]
        node[input neuron] (I-\name) at (0,-\y) {};

   \path[yshift=0.6cm,xshift=-2cm]
            node[hidden neuron, pin={[pin edge={draw=none}, pin distance = 0.005cm, xshift=-0.3cm, yshift=-0.2cm]above:${q_1}^{(1)}$}]  (H1-1) at (\layersep,-1 cm) {};
    
    \foreach \name / \y in {4,7}
        \path[yshift=0.6cm,xshift=-2cm]
            node[hidden neuron] (H1-\name) at (\layersep,-\y cm) {};

    \foreach \source in {1,...,5}
        \foreach \dest in {1,4,7}
            \path (I-\source) edge (H1-\dest);

 	\foreach \name / \y in {1,4,7}
        \path[yshift=0.6cm, xshift = 0.5cm]
            node[hidden neuron] (H2-\name) at (\layersep,-\y cm) {};
            

      \path[yshift=0.6cm,xshift=0.5cm]
             node[hidden neuron, pin={[pin edge={draw=none}, pin distance = 0.005cm, xshift=+0.6cm, yshift=+0.2cm]below:${q_H}^{(L)}$}]  (H2-7) at (\layersep,-7 cm) {};

	\path (H1-1) -- node[]{{\rotatebox{-90}\ldots}} (H1-4);
	\path (H1-4) -- node[]{{\rotatebox{-90}\ldots}} (H1-7);
    \path (H2-1) -- node[]{{\rotatebox{-90}\ldots}} (H2-4);
	\path (H2-4) -- node[]{{\rotatebox{-90}\ldots}} (H2-7);
    \path (H1-1) -- node[]{\ldots} (H2-1);
	\path (H1-4) -- node[]{\ldots} (H2-4);
	\path (H1-7) -- node[]{\ldots} (H2-7);
	\path (H1-4) -- node[]{{\rotatebox{+45}\ldots}} (H2-1);
	\path (H1-4) -- node[]{{\rotatebox{-45}\ldots}} (H2-7);
    

    \node[output neuron, fill=red!40, pin={[pin edge={draw=none}, pin distance = 0.005cm, xshift=+0.2cm, yshift=-0.1cm]above:$r_1$}] (r)  at (2*\layersep,-2 cm) {};
    \node[] (s)  at (2*\layersep,-3.5 cm) {\rotatebox{-90}{\ldots}};
    \node[output neuron, fill=blue!40, pin={[pin edge={draw=none}, pin distance = 0.005cm, xshift=+0.2cm, yshift=+0.1cm]below:$r_I$}] (t)  at (2*\layersep,-5 cm) {};

    \foreach \source in {1,4,7}
        \path (H2-\source) edge (r);
    \foreach \source in {1,4,7}
        \path (H2-\source) edge (s);
     \foreach \source in {1,4,7}
        \path (H2-\source) edge (t);
        
\path (r) edge (V1);
\path (s) edge (V2);
\path (t) edge (V3);

\end{scope}

\end{tikzpicture}
\caption{\label{fig:hyb} L-MNL model architecture. On the top, we have the $I$ class generalization of a linear-in-parameter MNL model, as depicted in Figure \ref{fig:CNN}. At the bottom, we have a deep neural network (\textit{i.e.}, multilayer and fully connected) that enables us to obtain the representation learning term $r_i$. The terms from each part are added together defining the new systematic function of Equation (\ref{eq:systematicutility}).}
\end{figure}
\end{center}

\subsection{ {L-MNL nested generalization}} \label{sec:nests}

The Nested Logit formulation has been first introduced by \cite{williams1977formation}. It relaxes the \textit{i.i.d.} assumption of MNL random terms in the case where alternatives may have correlation among each other in the decision making process. They are gathered together in nests which then affects their final probability such that: 
\begin{align}\label{eq:nested_proba}
    P_n(i|\mathcal{C}) &= P_n(i|m)\cdot P_n(m|\mathcal{C}) \notag \\
    &= \frac{e^{\mu_m V_{in}}}{\sum_{j\in \mathcal{C}_m} e^{\mu_m V_{jn}}} \cdot   \frac{\exp\left(\frac{\mu}{\mu_m}\ln\sum_{p\in\mathcal{C}_m} e^{V_{pn}}\right)}{\sum_{k=1}^M\exp\left( \frac{\mu}{\mu_k} \ln\sum_{p\in \mathcal{C}_k} e^{\mu_k V_{pn}}\right)}
\end{align}
 where $M$ is the total number of nests, $\mu_l$ is the factor for nest $l$ and $\mathcal{C}_l$ is the set of alternatives belonging to that nest.
This is the probability of choosing an alternative within the nest $m$ multiplied by the probability of choosing $m$ among all nests. 

To add a representation learning term to the Nested Logit, one must only change the final function of L-MNL seen in Equation (\ref{eq:softmax}) and adapt it to Equation (\ref{eq:nested_proba}). This is done by removing the commonly used softmax layer seen in Figure \ref{fig:hyb} for a custom-built network with added trainable weights for the nest factors.{ The new architecture is depicted in Figure \ref{fig:nested_loss} of Appendix E}. Further changes to the network would allow us to easily model Cross-Nested Logit (\cite{vovsha1997application}) and Generalized Extreme Value models (\cite{mcfadden1978modeling}) with an added representation term. 
\color{black}

\section{Experiments}\label{sec:experiments}

Our experiments demonstrate how our L-MNL and L-NL models increase the \textit{predictability} of the MNL and NL models respectively while keeping their \textit{interpretability}. The former,  \textit{predictability}, is directly quantified in terms of likelihood using two real-world datasets. The latter \textit{interpretability}, however, is more challenging to assess. To do so, we define interpretability as the ability of the model to have both a quantifiable uncertainty of its parameters as well as an understandable or meaningful association with its variables. This latter point usually translates as the ability to mathematically derive and interpret the impact, or elasticity, of a variable. We will use  synthetic and semi-synthetic data for which the true parameters are known to assess the quality of the estimates and the interpretability of every model.  
 
\subsection{Benchmarking models} \label{sec:benchmark_models}
We want to compare our models (L-MNL and L-NL) against key previous works.
As our approach is both general and flexible, we retrieve previous works with appropriate modeling decisions on the $\mathcal{X}$, $\mathcal{Q}$ input sets and $\bm{f}$, $\bm{r}$ functions of Equation (\ref{eq:ourutility}).

Essentially, we compare the following models:

\begin{itemize}
\item Logit($\mathcal{X}$): The standard MNL model with linear-in-parameter specification, and no learning component. The variables $x \in \mathcal{X}$ enter the linear utility specification and $\mathcal{Q} = \emptyset$. \red{The model has $|\mathcal{X}|$ parameters.}
\item {DNN}($n,\mathcal{Q}$): {A Dense Neural Network for every alternative with softmax loss.} This is the NN-MNL model proposed by \cite{hruschka2004empirical} {and based on \cite{bentz2000neural} with $n$ neurons in the hidden layer ($H=n$) and a single layer deep.} The variables $x \in \mathcal{Q}$ are the input features of the neural network. There is no variable entering the linear utility specification ($\mathcal{X}=\emptyset$). \red{The model has around $ n(|\mathcal{Q}| + |\mathcal{C}| + 1)$ weights.}
\item \red{DNN\_L}($n,\mathcal{X}=\mathcal{Q}$): A modified version of the NN-MNL model proposed by \cite{hruschka2004empirical}. The variables $x \in \mathcal{X} (= \mathcal{Q}$) enter both the linear utility specification and the neural network. \red{The latter can be seen as finding the residual from the linear input of $\mathcal{X}$ and the overall architecture is close to the concept of the more well known residual building block from ResNet (\cite{he2016deep})}. \red{The model has in the order of $(n+1)(|\mathcal{X}| + |\mathcal{C}|)$ weights.} 
\item L-MNL($n,\mathcal{X},\mathcal{Q}$): Our learning logit model with $n$ neurons in the hidden layer ($H=n$). Part of the variables $x \in \mathcal{X}$ enter the linear utility specification and the remaining variables $q \in \mathcal{Q}$ are given as input features to the neural network. This separation satisfies the interpretability condition (Equation \ref{eq:criticalassumption}) for all $x$. The function $\bm{r}(\mathcal{Q})$ is defined by Equation (\ref{eq:NN_component}) {where we limit the complexity to a single layer ($L=1$). \red{Therefore, the model has a total of $|\mathcal{X}|$ interpretable parameters and $n(|\mathcal{Q}| + |\mathcal{C}|)$ weights.} We remind the reader that this architecture for the representation term is for the sake of example and more complex networks or methods depend on the datasets at hand and are open for future research.}
\end{itemize}

{Unless specified otherwise, all models have run for 200 epochs with an Adam optimizer (\cite{kingma2014adam}) running on default parameters from Keras python deep learning library (\cite{chollet2015keras}). Every model with a Neural Network has a $20\%$ dropout regularizer following the DNN layer. The parameters in every model are initialized by the default random initialization of the used library, and no seed is set. Moreover, in our current framework, no notable difference in learning time between models are observed.} 


\subsection{Synthetic data} \label{sec:synthetic}
{We now use synthetic data to better analyze the performance of our L-MNL model in terms of both prediction performance and estimates accuracy.} We start by describing how we generated synthetic data. Then, we perform Monte Carlo experiments to compare all benchmarking models on parameters estimation. Then, a scan on increasing number of neurons is introduced to analyze the impact of the NN architecture better. {Finally, we investigate the case of strongly correlated variables between input sets $\mathcal{X}$ and $\mathcal{Q}$.} {Additional experiments to study the particular effects of complexly correlated variables, missing variables, or sequential versus joint optimization strategies can be found in \ref{sec:Synth_Annex}.} To study the limits of the hybrid model, a special case study of very noisy and highly non-linear utility specifications using semi-synthetic data can be found in \ref{sec:semi_synthetic}. { For all experiments, all benchmarking models remain the same as in the previous Section (\ref{sec:benchmark_models}).}

\subsubsection{Data generation}\label{sec:datageneration}
Synthetic data provides a controlled environment for analyzing our L-MNL model. To generate our synthetic data, we consider a simple binary choice model where an individual $n$ has to choose between two options (\textit{e.g.}, choosing to use public transportation to go to work as opposed to taking an alternative mode). {Following the generation process of \cite{guevara2015critical}, we obtain the following utility specifications for $i=1,2$:
}

\begin{equation} \label{binarylogit}
  U_{in} =  V_{in} + \varepsilon_{in},
 \end{equation}
 with 
 
 \begin{equation} \label{eq:binarylogit_systematic}
{
V_{in} = \underbrace{\beta_p \cdot p_{in} + \beta_a\cdot a_{in} + \beta_b\cdot b_{in}}_{\textit{known relation}} \hspace{0.5cm} + \underbrace{\beta_{qc}\cdot q_{in} c_{in} }_{\textit{unknown interactions}}
}
 \end{equation} 
 {
\begin{align}
  & & p_{in} & =  5+ z_{in} + 0.03wz_{in} + \varepsilon_{pin} & \\
  & &  q_{in} & = 2 h_{in} +k_{in} + \varepsilon_{qin} & \\
  & & k_{in} & = h_{in} + \varepsilon_{kin} & \text{for  } i= 1,2
\end{align}
\noindent where $\beta_x$ stand for parameters, and all variables $a_{in}$, $b_{in}$, $c_{in}$, $z_{in}$, $wz_{in}$, $h_{in}$ and error terms $\varepsilon_{pin}$, $\varepsilon_{qin}$, $\varepsilon_{pin}$ follow a uniform distribution.
}
For the following experiments we consider the non-linear interactions among variables to represent unknown and undiscovered causalities during the modeling phase. Our data generation process slightly differs from  \cite{guevara2015critical} on two points:
first of all, $p_{in}$ is not correlated with $q_{in}$ as this is kept for an experiment by itself in Section \ref{sec:correlation} and \ref{sec:guevara}. Secondly, we have added an interaction term, $c_{in}$ to allow a simulated situation where the modeler would misspecify the utility due to an undiscovered interaction. To avoid too large difference in magnitude order among our variables as well as the interaction term, we generated all variables as:
\begin{equation}
    x_{in} \sim \mathcal{U}([-1,1])
\end{equation}

Under the classical logit assumption that $\varepsilon_{in} \overset{\text{i.i.d.}}{\sim} EV(0,1)$ and given fixed values of parameter estimates $\bm{\beta}$ in the utility specification, the choice probabilities $P_{in}(\bm{x_n})$ are known, and the synthetic choice $y_{in}$ can then be seen as a Bernoulli random variable that takes the value 1 with probability $P_{in}(\bm{x_n})$ and 0 with probability $1-P_{in}(\bm{x_n})$. Formally, we assume that
{
\begin{align} \label{syntheticchoice}
& & y_{1n} &\sim \operatorname{Bern} \left({P_{1n}(\bm{x_n})}\right), &  \\
& & y_{2n} &= 1-y_{1n} &  \forall n =1,...N.
 \end{align}
}
Using synthetic data allows us to compare our L-MNL model with standard choice models, not only in terms of prediction performance, but also in accuracy in parameter estimation since we know the true value of parameter estimates $\bm{\beta}$, \textit{i.e.}, the true linear and non-linear dependencies between the variables. 

\subsubsection{Monte Carlo experiment}\label{montecarlo}
We rely on Monte Carlo simulations to investigate how well each model performs in both prediction and parameter estimation of Equation (\ref{eq:binarylogit_systematic}). \red{To do so, we select a true utility specification with $\beta_p=-1$, $\beta_a = 0.5$, $\beta_b$ = 0.5, $\beta_{qc}=1$. For each experiment, we generate $1,000$ synthetic individual observations for the training set, and $200$ more for the testing set. The following results are obtained by performing $100$ of such experiments.}\\

Our new choice model is defined as
\begin{itemize}
\item L-MNL($\red{25},\mathcal{X},\mathcal{Q}$), with $\mathcal{X} = \{\bm{p},\bm{a},\bm{b}\}$, and $\mathcal{Q}=\{\bm{q}, \bm{c}\}$,
\end{itemize}
and is compared with the following three benchmarking models and one reference model*\footnote{The reference model is considered differently than the benchmarks as it breaks the assumption that the modeler does not know the ground truth.}:
\begin{itemize}
\item Logit($\mathcal{X}_1$), with $\mathcal{X}_1 = \{\bm{p},\bm{a},\bm{b},\bm{q}, \bm{c}\}$,
\item DNN($\red{25},\mathcal{Q}$), with $\mathcal{Q} =  \{\bm{p},\bm{a},\bm{b},\bm{q}, \bm{c}\}$,
\item \red{DNN\_L($25,\mathcal{X}=\mathcal{Q}$), with $\mathcal{X} = \mathcal{Q} = \{\bm{p},\bm{a},\bm{b},\bm{q}, \bm{c}\}$,}
\item *Logit($\mathcal{X}_{true}$), with $\mathcal{X}_{true} = \{\bm{p},\bm{a},\bm{b},\bm{qc}\}$.
\end{itemize}

The log-likelihood results from Monte Carlo experiments can be seen in Table \ref{tab:montecarlo_LL}. As expected, the best models in terms of predictive performance are the neural network-based choice models. Among them, despite overfitting on the train set, our L-MNL gives the best general representation of the data, achieving the best predictive performance in the test set (with a LL of -97 compared to -123 for the standard MNL model).   

\begin{table}[tb]
\color{black}
\centering
\caption{Monte Carlo average log-likelihood ($\bar{LL}$) and standard deviation ($s.d.(LL)$) for the different models. Based on the test set value, we conclude that our L-MNL learns the best general representation.  An MNL model with true utility specification is given as a reference. Average of accuracies }
\label{tab:montecarlo_LL}

\begin{tabular}{l|cc|cc|cc|c}

\multicolumn{1}{c}{\textbf{Model}} & \multicolumn{2}{c}{Train set} & 
\multicolumn{2}{c}{Test set} & \multicolumn{2}{c}{Accuracies [\%]} & $\rho^2_{test}$\\

 &$\overline{LL}$ & $s.d.(LL)$&$\overline{LL}$ & $s.d.(LL)$ &$\overline{Acc}_{train}$ & $\overline{Acc}_{test}$ \\ 
\hline
*Logit($\mathcal{X}_{true}$) & \textit{-459} & \textit{15}   & \textit{-94} &\textit{8}  &\textit{78}  &\textit{77} & \textit{0.32}\\[1pt]
\hline
Logit($\mathcal{X}_1$) & -604& 11 & -123& 6 &67 &66 & 0.11\\[1pt]
\hline
DNN($25,\mathcal{Q}$) & -363 & 21 & -112& 13 &84&74 & 0.19\\[1pt]
\hline
DNN\_L($25,\mathcal{X}=\mathcal{Q}$) & -367 & 18 & -108& 12 & 83& 75 & 0.22\\[1pt]
\hline
L-MNL($25,\mathcal{X},\mathcal{Q}$) & -429 & 16 & $\bm{-97}$& $\bm{8}$ &80 & $\bm{76}$ & $\bm{0.30}$\\[1pt]
\end{tabular}
\end{table}

In order to evaluate the models in terms of accuracy in interpretable parameter estimation, we define the relative errors $e_{\beta}$ and $e_{\beta_i/\beta_j}$ as
\begin{align}
e_{\beta} &=  \left\lvert \frac{\beta -  \widehat{\beta}}{\beta} \right\rvert,\\
e_{\beta_i/\beta_j} &= \left\lvert \frac{e_{\beta_i}-e_{\beta_j}}{1-e_{\beta_j}} \right\rvert.
\end{align}
The relative error results\footnote{Note that the DNN($25,\mathcal{Q}$) benchmarking model does not appear in Table \ref{tab:montecarlo_errors} since no variable enters the linear utility specification.} from Monte Carlo experiments can be seen in Table \ref{tab:montecarlo_errors}. \red{As the beta parameters of the model have different values, we normalize the performance by reporting the average relative error of the estimates for every model, as well as the standard deviation of all relative errors.} We see that our L-MNL greatly outperforms every model in the ability to recover the true parameter values with a relative error smaller than $1\%$ away from the true model. It is worth noting that for the ratios of parameters, the MNL models perform second best, although they have high relative errors at parameters level. This phenomenon has been investigated in a few papers (\textit{e.g.}, \cite{lee1982specification,cramer2007robustness}) where it has been shown that even when utility is misspecified, the MNL model is still able to retrieve good ratios between estimates. The hybrid models with $\mathcal{X}=\mathcal{Q}$, on the other hand, generate high errors in parameter estimates. It indicates that the NN component is also partially learning linear dependencies of $\bm{p}$ or $\bm{a}$, and prevents the linear function $f$, or the $\beta$ parameters, to reach a minimum as good as our L-MNL. {This is a direct consequence from Equation (\ref{eq:elasticity}) as the impact of the variable is shared between both data-driven and knowledge-driven terms.}
\begin{table}[tb]
\color{black}
\centering
\caption{Monte Carlo relative errors for the different models in [$\%$] with $\overline{e}$ the average relative error, $s.d.$ its standard deviation and $\beta_p$, $\beta_a$ are from Equation (\ref{eq:binarylogit_systematic}). }
\label{tab:montecarlo_errors}
{
\begin{tabular}{l|cc|cc|cc}
\textbf{Model} & \textbf{$\overline{e}_{\beta_p}$} & $s.d.(e_{\beta_p})$ & \textbf{$\overline{e}_{\beta_a}$} & $s.d.(e_{\beta_a})$ & \textbf{$\overline{e}_{\beta_p/\beta_a}$} & $s.d.(e_{\beta_p/\beta_a})$\\
\hline
*Logit($\mathcal{X}_{true}$) & \textit{6.4} & \pmtab{\textit{4.9}} & \textit{14.4} &\pmtab{\textit{10.7}} & \textit{10.8}&\pmtabend{\textit{9.7}}\\
\hline
Logit($\mathcal{X}_1$) &26.7 &\pmtab{6.2} & 26.7&\pmtab{14.7}& 15.5& \pmtabend{12.4}\\[1pt]
\hline
{DNN\_L}($25,\mathcal{X}=\mathcal{Q}$) &60 &\pmtab{32.4} &74&\pmtab{54}&460& \pmtabend{166} \\[1pt]
\hline
L-MNL($25,\mathcal{X},\mathcal{Q}$) & $\bm{7.1}$& \pmtabbf{$\bm{5.1}$} & $\bm{15.2}$& \pmtabbf{$\bm{11.7}$}& $\bm{11.3}$& \pmtabendbf{$\bm{10.3}$}\\[1pt]
\hline
\end{tabular}
}
\end{table}

Finally, we conduct hypothesis testing to determine whether the estimated parameters are statistically different from the true ones. Formally, we consider the following null and alternative hypotheses: $H_0: \widehat{\beta} = \beta$, and $H_1: \widehat{\beta} \neq \beta${, where the null hypothesis is rejected for a t-test above 1.96, which corresponds to a 5$\%$ confidence interval. The standard deviation of the estimates in $\mathcal{X}$ for the t-tests are obtained via Hessian approximation\footnote{\color{black}Note that profile likelihood approaches such as Hessian approximation create confidence intervals that fail to have the desired frequentist coverage properties. See \cite{brunero} for further discussion on that topic.}}.

The results are shown in Table \ref{tab:montecarlo_testing}. We see that the coefficients alone are almost always statistically different from the true ones for every model except L-MNL. The NN component of our model fixes the estimators by learning only on the data that are included in the linear component. The ratio of parameters is well retrieved with both MNL and L-MNL models. Concerning the residual block model, we see that its NN component compromises the ratios between parameters. Indeed, as seen in Equation (\ref{eq:criticalassumption}), part of the Neural Network has overlapped over the estimates of $\beta_p$ and $\beta_a$, thus breaking both the estimates and their ratio.

To better illustrate the origin of ratio discrepancy {of variables $x_{in}\in\mathcal{X}$ when $\frac{\partial r_{in}}{\partial x_{in}} \neq 0$ } in hybrid models, we {have shown in Section \ref{sec:correlation}} that we retrieve the same biasing effect on our L-MNL when we make use of highly correlated data between both sets. 

\begin{table}[tb]
\color{black}
\centering
\caption{Monte Carlo hypothesis testing for the different models, for $\beta_p$ and $\beta_a$ taken separately and for their ratio. Parameters $\beta_p$, $\beta_a$ are from Equation (\ref{eq:binarylogit_systematic}).}
\label{tab:montecarlo_testing}
{
\begin{tabular}{l|c|c}
\multicolumn{1}{c}{\textbf{Model}} & \multicolumn{2}{c}{\textbf{\% of experiments not rejecting $H_0$}} \\
 & $\beta_p$ and $\beta_a$ & $\beta_p/\beta_a$ \\
\hline 
*Logit($\mathcal{X}_{true}$) & \textit{97.5} & \textit{94} \\
\hline
Logit($\mathcal{X}_1$) & \hspace{35pt}34\hspace{35pt} & $\bm{97}$\\[1pt]
\hline
DNN\_L($25,\mathcal{X}=\mathcal{Q}$) & 25.5 & 37\\[1pt]
\hline
L-MNL($25,\mathcal{X},\mathcal{Q}$) & $\bm{95}$ & $\bm{94}$\\[1pt]
\hline
\end{tabular}
}
\end{table}

\subsubsection{Choice of neural network architecture} \label{sec:overfit}
\begin{figure}[t]
	\begin{center}
	\includegraphics[width=0.8\textwidth,angle=0]{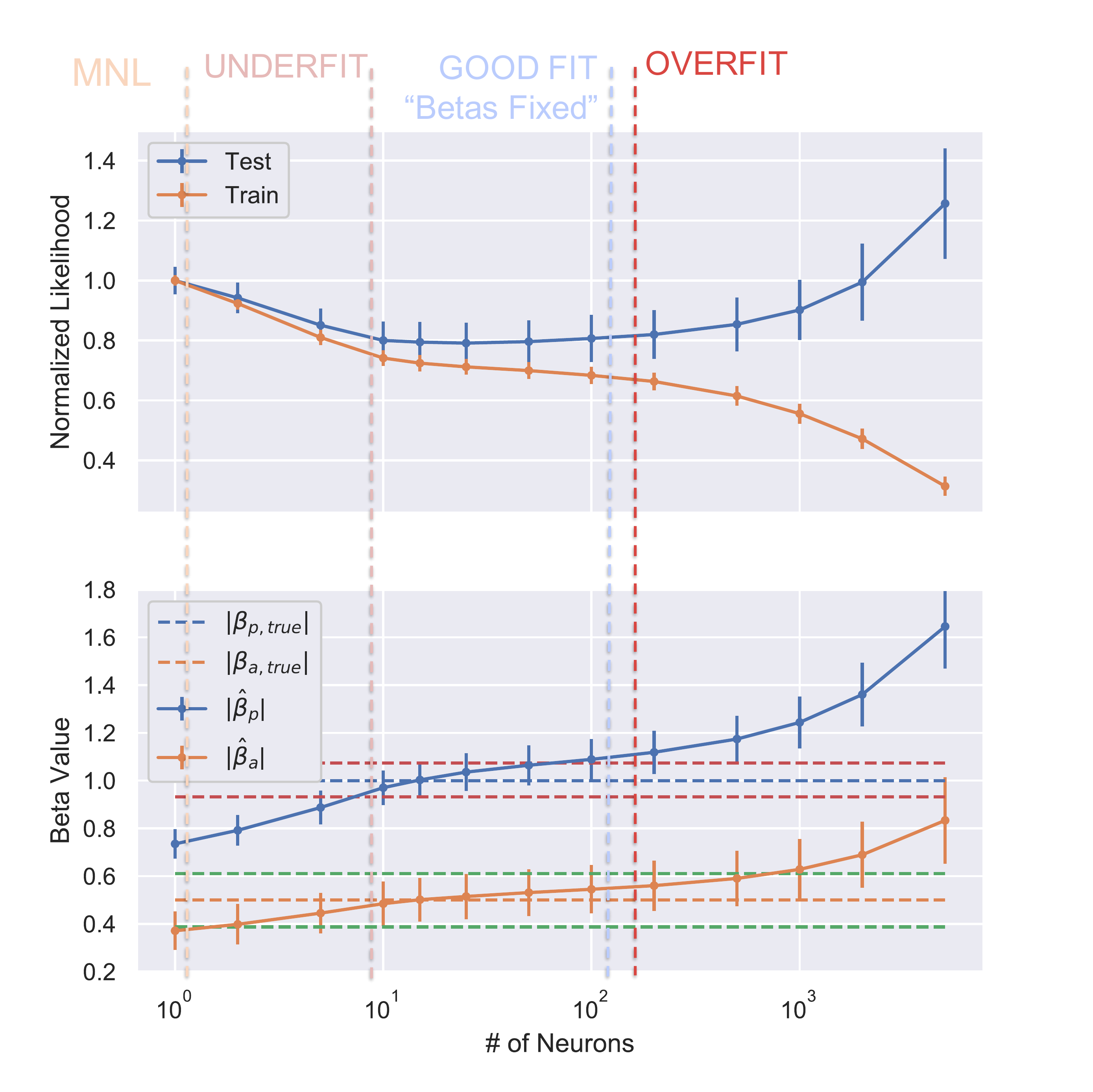}
	\end{center}
	\caption{ \label{fig:ToyLines} {Likelihood and values of parameter estimates for an increasing number of neurons in the hidden layer.\red{ Error bars show the standard deviation for 100 experiments. Red and Green lines show the standard deviation spread of the true model's parameter estimation.} Best results are obtained with $n\in[10,100]$ }}
\end{figure}

Choosing the structure of an NN requires one to have enough capacity to capture the underlying pattern of the data without overfitting. In the following, we gradually increase the size of the NN component of our L-MNL, spanning from an underfit of the non-linear \textit{truth} of Equation (\ref{eq:binarylogit_systematic}) to an overfit of the generated data. To illustrate these effects, we study the values of {$\beta_{p}$ and $\beta_{a}$}, as well as likelihoods of both train and test sets.\\
The results are depicted in Figure \ref{fig:ToyLines}, where we take L-MNL($n,\mathcal{X},\mathcal{Q}$) with a single layer L=1, and scan from $n=0$ neurons in the hidden layer to $n=5000$. We see that from $n=0$ neurons ($\equiv$MNL) to about $n\sim10$, the NN has not yet captured all the non-linearities of the original utility specification. This can be seen by the higher values in likelihood and how the parameter estimates have not yet reached their true minimum. From {$n=10$ to $n=100$}, the model reaches stable values where the linear terms are equivalent to their ground truth. {The model performs almost as well as the true model, depicted by the boundary lines, leading us to believe that the} NN component has successfully learned the non-linearities of the data. When we continue to increase the number of neurons, we see the effects of overfitting, namely a drop in train likelihood and an increase in test likelihood. We finally observe that the values of {$\beta_{p}$ and $\beta_{a}$} also start to suffer from the overfitting, as the L-MNL is no longer a general representation of the data, but has become specific to the training set. This experiment also shows that for equal optimal likelihood values, the change in architecture and non-convex property of the neural network does not significantly affect the $\beta$ estimates as it converges to the ground truth. {We can see that the standard deviation spread of the estimates over 100 experiments match with those of the true model also reported on the figure.} Although, is has been shown in \cite{draxler2018essentially} and \cite{garipov2018loss} that a neural network may have many different parameter estimates for similar valued optimum, it does not seem to be the case for our convex part of the model.

\subsubsection{Impact of strongly correlated variables}\label{sec:correlation}
Experimental results from Section \ref{montecarlo} suggest that having identical variables (or features) in both sets lead to poor parameter estimates because of the neural network's better ability to discover {correlation with the observations for all its variables. In this sense, Equation (\ref{eq:elasticity}) shows how a variable's impact will be shared between both knowledge-driven and data-driven terms if it belongs to both input sets.  }

Although our L-MNL requests the variables in the two sets to be different, strong correlation between the variables can remain an issue. To investigate the impact of correlated variables, we replace the original explanatory variable $q_{in}$ with a new explanatory variable $q'_{in}$ that is defined to be correlated to $p_{in}$. Formally, we define $q'_{in}$ as
\begin{equation}\label{eq:correlation}
q'_{in}= s\cdot p_{in} + \sqrt{1-s^2}\cdot q_{in},
\end{equation}
where $s\in [0,1]$ is the correlation coefficient.\\

Our new choice model is defined as
\begin{itemize}
\item L-MNL($100,\mathcal{X},\mathcal{Q}$), with $\mathcal{X} = \{\bm{p},\bm{a},\bm{b}\}$, and $\mathcal{Q}=\{\bm{q'},\bm{c}\}$,
\end{itemize}
and is compared with the following two benchmarking models:
\begin{itemize}
\item Logit($\mathcal{X}_1$), with $\mathcal{X}_1 = \{\bm{p},\bm{a},\bm{b},\bm{q'},\bm{c}\}$,
\item Logit($\mathcal{X}_2$), with $\mathcal{X}_2 = \{\bm{p},\bm{a},\bm{b}\}$.
\end{itemize}

\begin{figure}[t]
\centering
	\includegraphics[width=0.9\textwidth,angle=0]{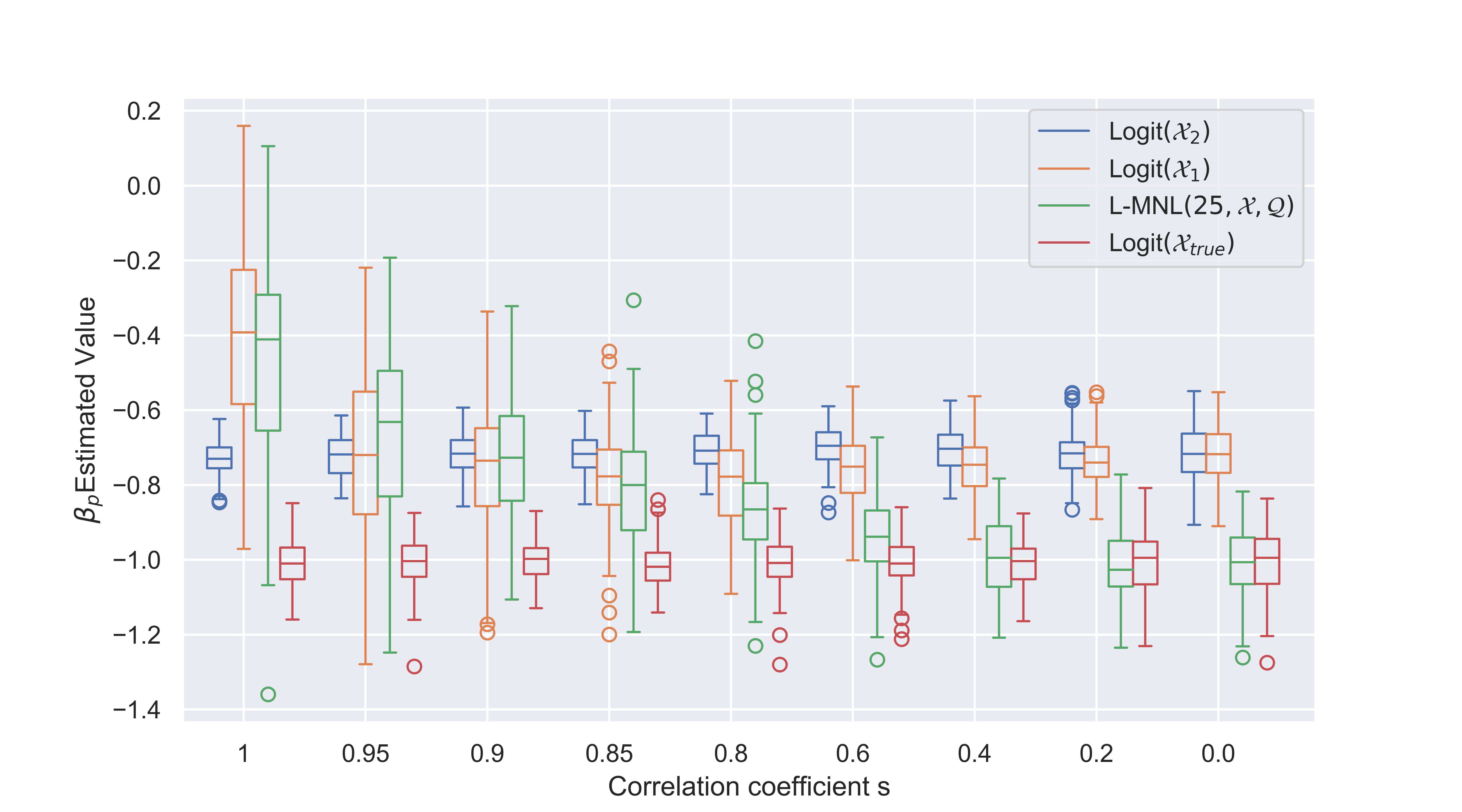}
\caption{ Impact of correlated variables on parameter estimates, where $\bm{p}\in\mathcal{X}$ is correlated to $\bm{q'}\in\mathcal{Q}$ (see Eq.\ref{eq:correlation}). We see in this case that the L-MNL correlation bias is smaller than the bias due to underfit for all $s\leq0.8$. For each coefficient, we performed 100 experiments.} \label{fig:barplot}
\end{figure}

The Monte Carlo mean relative errors for different levels of correlation are shown in Figure \ref{fig:barplot}.
\red{
Overall we can see two effects: bias with high variance due to the correlated variables (Logit($\mathcal{X}_2$), L-MNL), and bias due to misspecification (Logit($\mathcal{X}_1$), Logit($\mathcal{X}_2$)). For a correlation coefficient below $s \leq 0.8$ we can see the L-MNL has much better estimates with respect to the true model than the Logit models. For $s \leq 0.4$ the L-MNL is practically as efficient as the true model.}  
This is important as it suggests that in order to get insightful results, the modeler has to carefully check that the variables that enter as input features in the neural network are not the same or not too strongly correlated with the variables in the linear utility specification.  \\

Finally, we also show in \ref{sec:guevara}, that our new choice model does not suffer from complexly correlated variables between both sets $\mathcal{X}$ and $\mathcal{Q}$.

\subsection{A real case study: the Swissmetro dataset}\label{sec:Swissmetro}
For the following experiments, we use the openly available dataset Swissmetro (\cite{bierlaire2001acceptance}). The Swissmetro dataset 
consists of survey data collected in Switzerland during March 1998. The respondents provided information to analyze the impact of a new innovative transportation mode, represented by the Swissmetro, a revolutionary mag-lev underground system.  Each individual reported the chosen transportation mode choice for various trips, including the car, the train, or the Swissmetro. The original dataset contains 10,728 observations. We removed observations for which the information regarding the chosen alternative was missing (9 observations). For convenience, we also decided to discard the observations for which the three alternatives car, train, and Swissmetro were not all available (1,683 observations). Our dataset contains  9,036 observations. We split this initial dataset into a training set of 7,234 observations and a test set of 1,802 observations. 


\subsubsection{Models Comparison} \label{sec:Swissmetro_models}
In Table \ref{tab:swiss}, we present a detailed comparison of all the benchmarking models with respect to log-likelihood in the training and testing sets as well as the parameters estimates. 
Since the true values of parameter estimates are unknown, we compare the values obtained with our L-MNL to the ones obtained using the MNL model described in \cite{bierlaire2001acceptance}, whose linear-in-parameter utility specification is described in Table \ref{tab:utility_functions} and variables described in Table \ref{datasemisyntethic}.  The results of parameter optimization for this MNL model are shown in Table \ref{tab:swiss}\footnote{It is worth noting that the values of estimates that we obtain slightly differ from the ones depicted in the original paper. This is normal since the original Swissmetro dataset is different from the training dataset that we used.}.

 \begin{table}[tb]
  \begin{center}
    \caption{Swissmetro benchmark utility function, from \cite{bierlaire2001acceptance} \label{tab:utility_functions}}
    \label{tab:utility}
    \begin{tabular}{|ll|lll|} 
      \hline
 & & \multicolumn{3}{c|}{Alternative}   \\
    \cline{3-5}
	Variables & & Car & Train & Swissmetro \\ 
    \hline
	ASC & Constant & Car-Const &  & SM-Const\\
	TT & Travel Time & B-Time & B-Time & B-Time\\
	Cost & Travel Cost &  B-Cost & B-Cost & B-Cost \\
	Freq & Frequency &  & B-Freq & B-Freq \\
	GA & Annual Pass &  & B-GA & B-GA \\
	Age & Age in classes &  & B-Age & \\
	Luggage & Pieces of luggage & B-Luggage &  & \\
	Seats & Airline seating & & & B-Seats\\
	\hline
    \end{tabular}
  \end{center}
\end{table}

  \begin{table}[tb]
 \begin{center}
 \small
 \caption{\blue{Variables in the Swissmetro dataset used for modeled component of the utility specification} \label{datasemisyntethic}}
\begin{tabular}{|l|l|} 
     \hline
\textbf{Variable} & \textbf{Description} \\
\hline
TT & Door-to-door travel time in [minutes], scaled by 1/100. \\
Cost  & Travel cost in [CHF], scaled by 1/100.\\
Freq & Transportation headway in [minutes]\\
GA &  Binary variable indicating annual pass holders (=1).\\
Age &  IntegeEs variable scaled with the traveler's age.\\
Luggage & Integer variable scaled with amount of luggage during travel.\\
Seats &  Binary variable for special seats configuration in Swissmetro (=1).\\
\hline
    \end{tabular}
    \end{center}
\end{table}

\normalsize

  \begin{table}[tb]
 \begin{center}
 \small
 \caption{Unused variables in the Swissmetro dataset \label{tab:extra_features}}
\begin{tabular}{|l|l|} 
     \hline
\textbf{Variable} & \textbf{Description} \\
\hline
Purpose: & Integer variable indicating the trip purpose (business, leisure, etc. )\\
First : & Binary variable indicating if first class (=1) or not (=0)\\
Ticket: &  Integer variable indicating the ticket type (one-way, half-day, etc.)\\
Who: &  Integer variable indicating who is paying the ticket (self, employer, etc.) \\
Male: &  Binary variable indicating the traveler's gender (0 = female, 1 = male)\\
Income: &  Integer variable indicating the traveler's income per year.\\
Origin: &  Integer variable indicating the canton in which the travel begins.\\
Dest: &  Integer variable indicating the canton in which the travel ends.\\
\hline
    \end{tabular}
    \end{center}
\end{table}
\normalsize

We next estimate our new choice model L-MNL($100,\mathcal{X}_1,\mathcal{Q}_1$), where $\mathcal{X}_1$ contains the same variables as the ones included in the utility specification of \cite{bierlaire2001acceptance}, and $\mathcal{Q}_1$ contains all the unused variables in the dataset. These variables are described in Table \ref{tab:extra_features}. The results of this model are depicted in Table \ref{tab:swiss}. We see that adding the representation learning component in the utility specification significantly increases the log-likelihood, suggesting that these variables contain information that helps to explain travelers' choice. However, we also observe that several estimates are not statistically different from zero ({p}-value $>$ 0.05). We have concluded in Section \ref{montecarlo} {and further developed in Section \ref{sec:correlation}}, that only strong correlation between variables can lead to biased parameter estimates. However, the correlation among the variables has been investigated and does not seem to be the issue here (see Table \ref{tab:coeffcorrelation} in Appendix for the correlation among variables). Moreover, the DNN\_L(100,$\mathcal{X}_1 = \mathcal{Q}$) model (see Table \ref{tab:swiss}) has one-to-one correlation among variables in both sets and does not loose significance in its parameters. One explanation could be that the coefficients in L-MNL have lost their significance due to the neural network's ability to learn better {which variables and interactions are most correlated to the data.} Therefore, the significance of the same parameters in the initial MNL model can originate from a bias due to the model's underfit. Also, with many explanatory variables being omitted in the initial MNL model, it is worth noting that the model is more likely to be subject to endogeneity, \textit{i.e.,} correlation among the dependent variables and the error term. Endogeneity is a well-known cause of bias in parameter estimates and can cause bias in the parameter estimates of the initial MNL model.


Finally, in Table \ref{tab:swiss}, we estimate the new choice model L-MNL($100,\mathcal{X}_2,\mathcal{Q}_2$), where $\mathcal{X}_2$ contains only the variables needed to compute the Value of Time (VOT) and Value of Frequency (VOF), \textit{i.e.}, the variables $time$, $cost$, and $frequency$. All other variables constitute the set $\mathcal{Q}_2$ and are therefore given to the neural network. We observe that all parameter estimates are significant, while the log-likelihood has greatly increased compared to the standard MNL model. In other words, the representation term was able to get information from the previously non-significant variables in the linear specification. 

{Furthermore, as L-MNL has a bigger feature space than the aforementioned models, we have designed a dummy coded Multinomial Logit as another benchmark. The parameter space contains homogeneous $\beta_x$ for the variables in $\mathcal{X}_2$ and alternative specific $\beta_{iq}$ for all variables contained in $\mathcal{Q}$. The full feature space is written as $\mathcal{X}_{dum} = \mathcal{X}_2\cup\mathcal{Q}_2$ and the results can be seen in Table \ref{tab:swiss}. The big difference in final likelihoods with our L-MNL models further demonstrate that this dataset indeed contains complex functions and interactions among its variables which may be difficult to capture by the modeler.}\\

 We show a comparison of VOT and VOF for the different models in Table \ref{tab:Param}. \red{We have also added a Cross-Nested Logit (CNL), and a Triangular Panel Mixture (TPM) model to the results for comparison purposes, whose detailed parameter values can be found in \ref{sec:SM_extra_models} and for which the implementation was taken from Biogeme examples\footnote{{https://biogeme.epfl.ch/examples.html}}}. The difference in log-likelihood values between the L-MNL and the Logit($\mathcal{X}_1$) suggests that the MNL model suffers from underfitting that leads to significant ratio discrepancy. {A similar conclusion was derived only under strong non-linearities in the dataset, suggested in \ref{sec:semi_synthetic}}. The linear specification of the MNL model does not allow to capture the Swissmetro's non-linear dependencies among variables. \red{Moreover, we may observe for both added DCM methods, that their ratio values move towards those of L-MNL as their likelihood increases. This trend implies a better specification in their utility would allow them to reach even more similar values in ratios and likelihood with our new choice model. } 

\begin{figure}[t]
    \begin{center}
    \includegraphics[width=0.6\textwidth,angle=0]{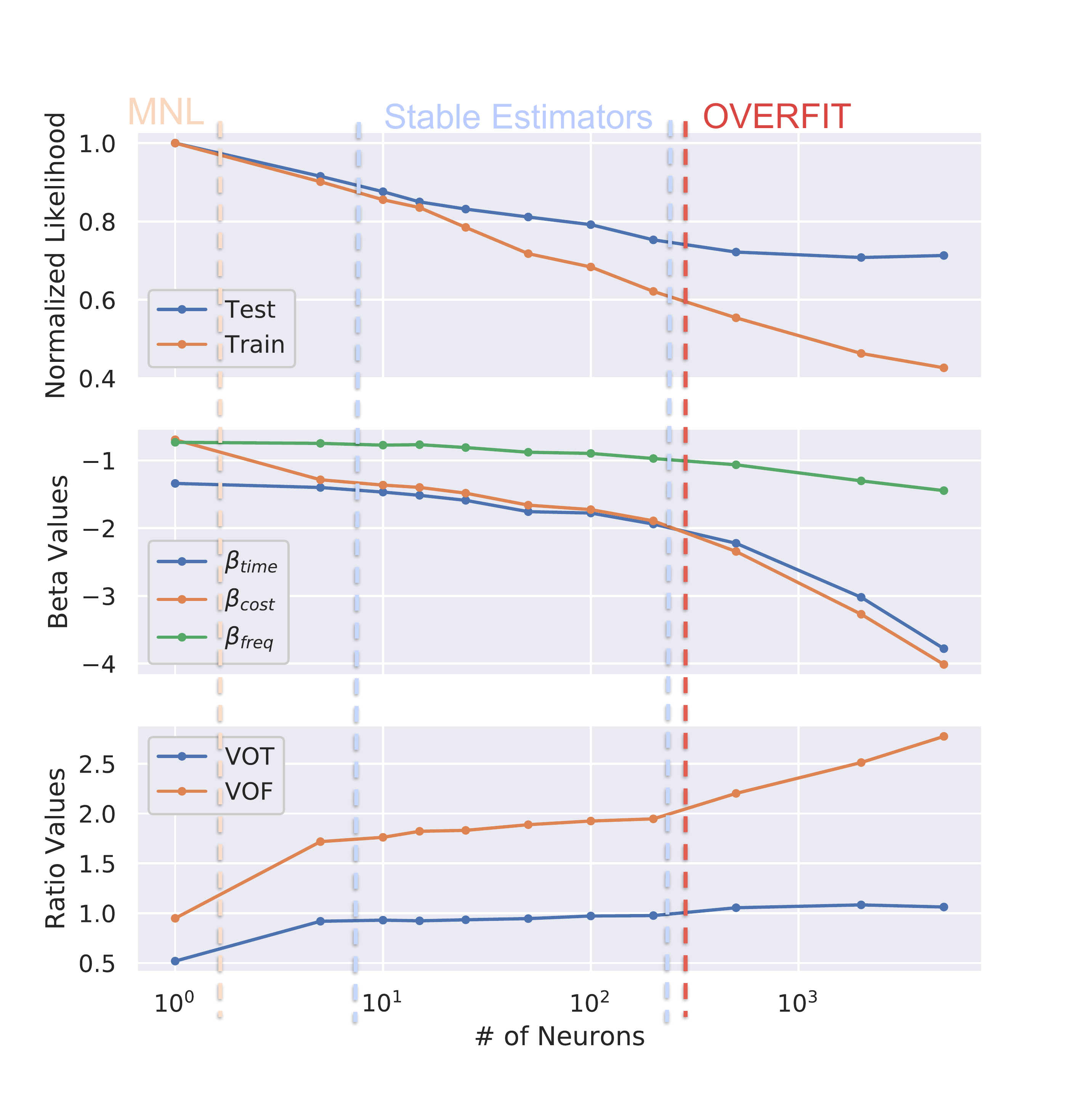}
    \end{center}
    \caption{  \label{fig:SwissLines} Scan of Likelihood and Beta values over number of neurons $n$ in the densely connected layer for L-MNL(100,$\mathcal{X}_2$,$\mathcal{Q}_2$). For $n\in[10,200]$ we have VOT$\approx 1$ and VOF$\approx1.9$. Over $n=200$, we have signs of overfit. }
\end{figure}

In support of these conclusions, we incrementally increase the size of the NN component. We observe in Figure \ref{fig:SwissLines} the parameter values with respect to the number of neurons in our neural network.
As for the synthetic data experiments in Section \ref{sec:overfit}, we see that stable ratios for VOT and VOF are obtained for a neural network having from 10 to 200 neurons. We see the MNL ratios for n=0 neuron highlights the underfit of the MNL model. An overfit effect is observed starting from 500 neurons, where the test likelihood no longer improves.


  \begin{table}[t]
  \caption{Comparison of log-likelihood and parameters estimates for different models with utility specification of \cite{bierlaire2001acceptance}. Number of observations = 7234. }\label{tab:swiss}

\begin{tabular}{llllll}
Model & Parameters & Estimates & Std errors & \textit{t}-stat & \textit{p}-value\\
\hline
 \multirow{9}{14em}{MNL\\ ${\rho}^2_{test} = 0.28$\\$\mathcal{L}(\hat{\beta}) = -5764$\\$\mathcal{L}_{test}(\hat{\beta}) = -1433$} & $ASC_{Car}$ & 1.08 & 0.162 & 6.67 & 0.00\\
 &$ASC_{SM}$ & 1.05 & 0.153 & 6.84 & 0.00\\
 &$\beta_{age}$ & 0.146 & 0.436 & 3.35 & 0.00\\
 &$\beta_{cost}$ & -0.695 & 0.0423 & -16.42 & 0.00\\
 &$\beta_{freq}$ & -0.733 & 0.1132 & -6.47 & 0.00\\
 &$\beta_{GA}$ & 1.54 & 0.167 & 9.24 & 0.00\\
 &$\beta_{luggage}$ & -0.114 & 0.0488 & -2.338 & 0.02\\
 &$\beta_{seats}$ & 0.432 & 0.115 & 3.76 & 0.00\\
&$\beta_{time}$ & -1.34 & 0.051 & -26.18 & 0.00\\
\hline
 \multirow{9}{14em}{L-MNL($100,\mathcal{X}_1,\mathcal{Q}_1$)\\${\rho}^2_{test} = 0.41$\\$\mathcal{L}(\hat{\beta}) =\bm{-4511}$\\$\mathcal{L}_{test}(\hat{\beta}) = \bm{-1181}$} & $ASC_{Car}$  & 0.106 & 0.174 & 0.61 & 0.54\\
 & $ASC_{SM}$ & 0.454 & 0.163 & 2.80 & 0.01\\ 
 &  $\beta_{age}$ & 0.390 & 0.045 & 8.63 & 0.00\\
  & $\beta_{cost}$& -1.378 & 0.048 & -28.45 & 0.00\\
  & $\beta_{freq}$ & -0.860 & 0.127 & -6.77 & 0.00\\
  & $\beta_{GA}$ & 0.214 & 0.194 & 1.10 & 0.27\\
  & $\beta_{luggage}$ & 0.116 & 0.0529 & 2.19 & 0.03 \\
  & $\beta_{seats}$ & 0.104 & 0.109 & 0.95 & 0.34\\
 & $\beta_{time}$  & -1.563 & 0.056 & -27.97 & 0.00\\
\hline
 \multirow{9}{14em}{DNN\_L(100,$\mathcal{X}_1=\mathcal{Q}$)\\${\rho}^2_{test} = 0.37$\\$\mathcal{L}(\hat{\beta}) = -4964$\\$\mathcal{L}_{test}(\hat{\beta}) = -1257$} & $ASC_{Car}$ & 0.365 & 0.165 & 3.61 & 0.00\\ 
 &$ASC_{SM}$  & 0.549 & 0.162 & 2.22 & 0.03\\
 &$\beta_{age}$ & 0.087 & 0.0423 & 2.07 & 0.04\\
 &$\beta_{cost}$& -0.897 & 0.046 & -19.46 & 0.00\\
 &$\beta_{freq}$ & -0.639 & 0.123 & -5.20 & 0.00\\
 &$\beta_{GA}$ & 1.40 & 0.172 & 8.15 & 0.10\\
 &$\beta_{luggage}$ & 0.186 & 0.0523& 3.52& 0.00\\
 &$\beta_{seats}$ & 0.233 & 0.102 & 2.29 & 0.02\\
&$\beta_{time}$  & -1.146 & 0.049 & -23.32 & 0.00\\
\hline
 \multirow{6}{14em}{\\Logit($\mathcal{X}_{dum}$) (all 41 inputs)\\${\rho}^2_{test} = 0.33$\\$\mathcal{L}(\hat{\beta}) = -5451$\\$\mathcal{L}_{test}(\hat{\beta}) = -1322$} & $\beta_{cost}$ & -1.062 & 0.059 & 18 & 0.00\\
 & $\beta_{freq}$ & -0.79 & 0.118 &  6.69 & 0.00\\
 & $\beta_{time}$ & -1.326 &  0.053 & 25.02 & 0.00\\
& ... &... &... &... &... \\
& & & & &\\
& & & & &\\
\hline
& & & & &\\
\multirow{5}{14em}{L-MNL($100,\mathcal{X}_2,\mathcal{Q}_2$)\\${\rho}^2_{test} = 0.44$\\$\mathcal{L}(\hat{\beta}) = \bm{-3895}$\\$\mathcal{L}_{test}(\hat{\beta}) = \bm{-1108}$} & $\beta_{cost}$ & -1.671  & 0.0523 & -31.94 & 0.00\\
& $\beta_{freq}$ & -0.865 & 0.0765 &  -11.30 & 0.00\\
& $\beta_{time}$ & -1.769 & 0.0389 & -45.4 & 0.00\\
& & & & &\\
& & & & &\\
\hline
\end{tabular}
 \end{table}


  \begin{table}[t]
  \begin{center}
  \caption{\normalsize{Parameter ratio comparison}}
  \label{tab:Param}
\begin{tabular}{c|cc|cc}
 {Model}   &  {Value of Time} &  {Value of Frequency} &  {Train Log-Likelihood} &  {Test Log-Likelihood} \\ \hline 
   {Logit($\mathcal{X}_1$)} &  {0.52} &  {0.95} &  {-5764}  & -1433 \\ 
Logit($\mathcal{X}_{dum}$)  & 0.80     & 1.34 & -5451  & -1322 \\ 
DNN\_L($\mathcal{X}_1$)   & 0.78     & 1.40 & -4964 & -1257 \\ 
L-MNL($\mathcal{X}_1$)  & 0.88   & 1.60  & -4511   & -1181\\ 
L-MNL($\mathcal{X}_2$) & 0.94     & 1.93 & $\bm{-3895}$  & $\bm{-1108}$   \\ 
\hline
\red{CNL($\mathcal{X}_1$)} & 0.59 & 1.52 & -5711 & -1415\\
\red{TPM($\mathcal{X}_1$)} & 0.72 & 1.59 & -4752 & -1350 

\end{tabular}
\end{center}
  \end{table}



\subsubsection{{Fixing artifact correlation between alternatives}}

In this experiment, we show results of a nested logit based on the baseline specification \cite{bierlaire2011projet} as well as the nested generalization of the L-MNL models seen in the previous experiment. \red{Moreover, we have added a model containing $\mathcal{X}_1$ and a smaller set $\mathcal{Q}_0=\{$Purpose, First, Male, Income$\}$ for illustrative purposes}. We may observe in Table \ref{tab:swiss_nests} the Log-Likelihoods for the standard models and their nested variants with the following nests: (1) \textit{known}: \{Car, Train\}, and (2) \textit{new}: \{Swissmetro\}.


  \begin{table}[t]
  \begin{center}
  \caption{\normalsize{Performance of Swissmetro models with added nest structure. Increase in fit lowers the need of nests.}}
  \label{tab:swiss_nests}
  
\begin{tabular}{l|cc|c|cc}
\multicolumn{1}{c}{} & 
\multicolumn{2}{c}{\makecell{\small{Standard Models}}} & \multicolumn{1}{c}{\multirowcell{2}[-1pt]{Nest Factor}} & \multicolumn{2}{c}{\makecell{\small{Nested Models}}} \\

	 & Train LL & Test LL & & Train LL& Test LL \\ 
\hline
 Logit($\mathcal{X}_1$) & -5764 & -1433 & \red{1.43} &  -5734 & -1413  \\
  L-MNL($\mathcal{X}_1,\mathcal{Q}_0$) & \red{-5324} & \red{-1348} & \red{1.37}   & \red{-5305} & \red{-1340} \\
 L-MNL($\mathcal{X}_1,\mathcal{Q}_1$) & \red{-4268} & -1173 & \red{1.01}   & \red{-4260} & \red{-1173} \\
 L-MNL($\mathcal{X}_2,\mathcal{Q}_2$) &  -3895 & 
 $\bm{-1108}$ & 1 &  \red{-3894} & $\bm{\red{-1108}}$  \\
 \hline
\end{tabular}
\end{center}
  \end{table}

  


The experiment shows that as we allow the representation term to find a data-driven result over an increasing amount of input, \textit{i.e.}, from \red{none to 
 $\mathcal{Q}_2$}, the nest value reduces itself to 1. If we relate this to the nested theory reminded in section \ref{sec:nests}, this means that the correlation between the choices in a nest --car together with train-- disappears as we better fit with a data-driven term. This suggests that the random terms were initially correlated between alternatives due to misspecification, but have become independent with the added representation term.  This interpretation is in line with the mathematical implications of Equation (\ref{eq:error_term}).

In further experiments, it would be possible to identify which variables are most active in the correlation between alternatives, thanks to ablation and elasticity studies. An example of such experiments can be seen in \ref{sec:sensitivity}. 
 \color{black}

\clearpage

\subsection{{Revealed Preference study: Optima dataset}} \label{sec:optima}
In contrast with the previous Swissmetro study, we investigate in this section a case much more difficult for representation learning due to the size limitation of gathered data. The project named Optima is a small survey collected in Switzerland between 2009 and 2010 where respondents filled up extensive information in the topic of mode choice, including time and cost of performed trips, socio-economic characteristics as well as opinions on statements and answers to "semi-open" questions as defined in \cite{glerum2014using}. More details on this particular dataset can be read in \cite{bierlaire2011projet} and \cite{glerum2014using}. To benchmark our results, we take the latest results discovered on this dataset, notably a well-specified base multinomial logit and one of its suggested extension, the Integrated Choice and Latent Variable (ICLV) model, from \cite{fernandez2016correcting}.

\subsubsection{Models description}
The multinomial logit specification from \cite{fernandez2016correcting} can be seen in Table \ref{tab:base_antolin}. We then compare two L-MNL models and two pure Neural Networks. The input sets for our first model is defined such that $\mathcal{X}_1$ contains the variables from the baseline specification, while $Q_1$ are those shown in Table \ref{tab:extra_features_optima}. \red{In this case it is interesting to note, that the straightforward interpretability condition for \textit{cost} in Section \ref{sec:modeling} is still satisfied, even though it interacts with \textit{income}$\in\mathcal{Q}$. Correlation between inputs of both sets can be found in \ref{sec:correlation_optima}. Based on experiments seen in Section \ref{sec:correlation} and \ref{sec:guevara}, this specification will not suffer bias from correlated variables.} The second L-MNL model has an input set $\mathcal{X}_2$, which contains only time, cost, and distance attributes, while the removed features were added into the second set to create $\mathcal{Q}_2$.
The Neural Networks have all features from both tables as inputs, but differ in size, the first having a single layer of $100$ neurons, just like the L-MNL models, and the second has a single layer of $30$ neurons.

We preprocess the data by removing any entry which does not have all choices available or unanswered variables we make use of. This gives us a total of 1376 answers, where 1089 are kept for the training set and 287 for the testing set. This is considered a tiny dataset for neural networks, which makes it a difficult task to avoid overfitting. Therefore, we have added a $30\%$ dropout layer and an $l_2$ regularizer of weight $\lambda = 0.5$ (\cite{krogh1992simple}).

 \begin{table}[tb]
  \begin{center}
  \caption{ {Base model specification from \cite{fernandez2016correcting} }}
    \label{tab:base_antolin}
    \begin{tabular}{|ll|lll|} 
        \hline
 & & \multicolumn{3}{c|}{Alternative}   \\
    \cline{3-5}
	Variables & & Public Transportation & Car & Slow modes \\ 
    \hline
	ASC & Constant & PT-Const & CAR-Const & \\
	TT & Travel Time [min] & B-Time-PT & B-Time-CAR & \\
	MCost & $\frac{\text{Marginal Cost}}{\text{Income}}$ &  B-MCost-PT & B-MCost-CAR &  \\
	Distance & Trip distance [km] &  & & B-Dist \\
	Work & Work related Trip &  & B-Work & \\
	French & French Speaking area &  & B-French & \\
	Student & Occupation is student & B-Student &  & \\
	Urban & Urban area & B-Urban & & \\
	NbChild & Number of Children &  & B-NbChild & \\
	NbCar & Number of Cars &  & B-NbCar &  \\
	NbBicy & Number of Bicycles &  & & B-NbBicy \\
	\hline
    \end{tabular}
  \end{center}
\end{table}

  \begin{table}[tb]
 \begin{center}
 \small
 \caption{ {Added variables for representation learning term in Optima dataset. All variables gave -1 for missing values.} \label{tab:extra_features_optima} }
\begin{tabular}{|l|l|} 
     \hline
\textbf{Variable} & \textbf{Description} \\
\hline
Age: &  Age of the respondent (in years) \\
HouseType : & 1 is individual house (or terraced house), 2 is apartment, 3 is independent room\\
Gender: & 1 is man, 2 is woman. -1 for missing value.   \\
Education: & Highest education achieved\footnote{More details at \url{https://biogeme.epfl.ch/data.html} \label{foot:optima_details} }. Categories from 1 to 8.  \\
FamilSitu: & Family situation\footref{foot:optima_details}.  Categories from 1 to 7. \\
ScaledIncome: &  Integer variable indicating the traveler's income per year.\\
OwnHouse: & Do you own the place where you are living? 1 is yes, 2 is no \\
MotherTongue: &  1 for german or swiss german, 2 for french, 3 for other, \\
SocioProfCat: & Socio-professional\footref{foot:optima_details} categories from 1 to 8.  \\
\hline
    \end{tabular}
    \end{center}
\end{table}

\subsubsection{Expert modeling as a regularizer}
\begin{figure}[t]
	\begin{center}
	\includegraphics[width=0.5\textwidth,angle=0]{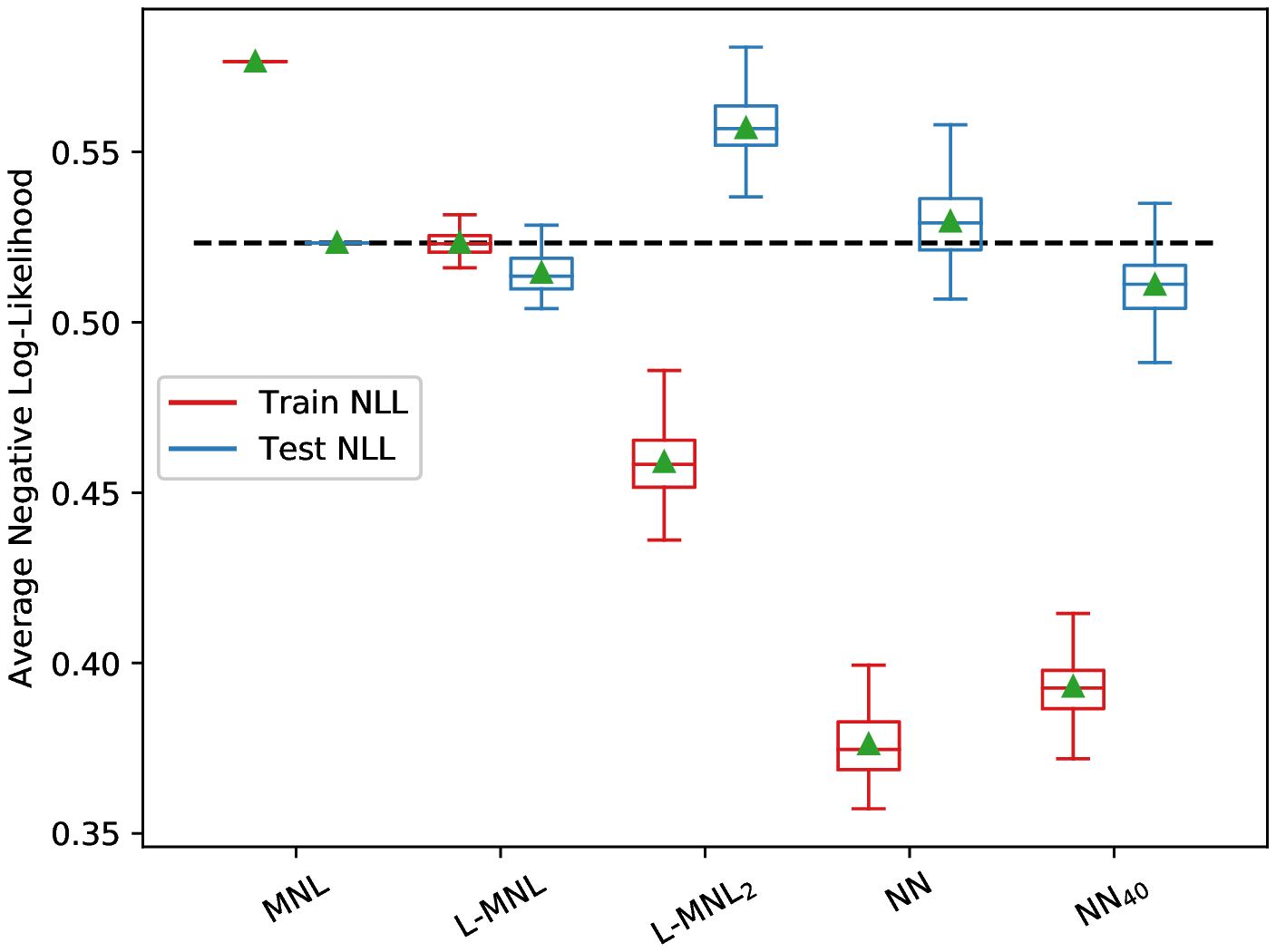}
	\end{center}
	\caption{ \label{fig:box_optima} Performance of multiple models on a small revealed preference dataset (Optima) for 100 minimization iterations. The first two models have expert modeling of the utility specification and generalize well. The last three have small or no modeling of the utility and show clear signs of overfitting as well as high variance in performance results. }
\end{figure}

The performances of the five models described above are shown in Figure \ref{fig:box_optima}, where the same minimization code was run 100 times for every model. The observed variation comes from the change in starting values, which ultimately brings to a slightly different optimum given the strong regularizers and fixed number of training cycles, \textit{i.e.}, $80$ epochs. 

We may observe that both MNL($\mathcal{X}_1$) and L-MNL($\mathcal{X}_1,\mathcal{Q}_1$) generalize well with an increase in performance for the second model. As opposed to the case seen in Swissmetro section \ref{sec:Swissmetro_models}, simplifying the $\mathcal{X}_1$ to the minimum of interest $\mathcal{X}_2$ allows the model to overfit and perform poorly. The same can be seen for a full neural network of $100$ neurons. If we try to reach better performances with a smaller neural network, including the high regularizers and low training cycles, we may observe the result of the last model, containing $40$ neurons. Although we still overfit on the training set, we have better generalization to the test set. However, the optimum has high variance, and parameter interpretation may only be recovered through elasticity plot studies, missing the straightforward interpretation of discrete choice model parameters.

\subsubsection{Benchmarking with ICLV}
In \cite{fernandez2016correcting}, the base logit is also compared to an ICLV model, where the latent variable CarLoving is added to the model. This latent variable is estimated with two Likert indicators (\cite{likert1932technique}):
\begin{itemize}
    \item $I_1$: It is difficult to take the public transport when I travel with
my children.
    \item $I_2$: With my car I can go wherever and whenever.
\end{itemize}
and the following specifications: 
\begin{equation}
t_{car} \cdot I_i = \alpha_i + \lambda_i \cdot CarLoving\cdot t_{car} + \omega_i
\end{equation}
where $\omega_i \sim \mathcal{N}(0,\sigma_i)$ is the random term, $\alpha_i$, $\lambda_i$, $\sigma_i$ are parameters to be estimated 
and: 
\begin{equation}
    CarLoving = \eta_{car} + \omega
\end{equation}
with $\eta_{car}$ and $\sigma$ being the estimated parameters and $\omega \sim \mathcal{N}(0,\sigma)$ the random term. 

Further details are described in \cite{fernandez2016correcting}. Given that the full loss function from an ICLV model differs from that of the base logit or MNL, we may only compare models by observing their accuracy on a test set. The implementation of the new model was done in biogeme (\cite{bierlaire2003biogeme}) and the results can be seen in \red{Table} \ref{tab:Optima_accuracy}. We may observe that the ICLV model performs better than MNL while having the same initial feature space. A stronger increase in accuracy could be expected with added latent variables or more complex structural equations. For the models with a neural network, we have reported the average accuracy of all iterations. As seen from the loss in Figure \ref{fig:box_optima}, the Neural Network ($n=40$) performs better than the L-MNL on average while slightly overfitting the training set. However, the obtained model does not contain straightforward interpretable parameters, has higher variance in performance, and required more efforts in fine-tuning for optimal performance. In the DCM sense, the most useful models would be L-MNL with the full expert specification and the ICLV model. Adding a representation term does not limit itself to MNL or its nested generalizations. We believe ICLV architectures may also benefit from a data-driven counterpart and we consider this for future research. 

  \begin{table}[h]
  \begin{center}
  \caption{\normalsize{Models accuracy on training and testing sets of Optima}}
  \label{tab:Optima_accuracy}
  
\begin{tabular}{l|cccc}
	Model & Logit($\mathcal{X}_1$) & L-MNL($\mathcal{X}_1$ ,$\mathcal{Q}_1$) & NN$_{40}$($\mathcal{X}\cup\mathcal{Q}$) & ICLV($\mathcal{X}_1$) \\
\hline
    Accuracy Train [$\%$] &  76.8   & 80.4 & 86.1 &  80.0  \\
    Accuracy Test [$\%$] & 76.7 & 79.2 & 81.3 & 77.7\\
 \hline
\end{tabular}
\end{center}
  \end{table}

In conclusion, when working with small datasets, Discrete Choice Modeling may effectively generalize very well when a good specification is found. The effect of the learning term may show how well our model is specified, \textit{i.e.}, a low increase in performance would signify a well-specified utility. Lastly, for the machine learning community, this may also encourage to use expert modeling of a utility specification as an efficient regularizer for neural networks when working with small datasets.

\section{Future Directions}

Our proposed framework paves the way to several avenues of research. One main direction is the new possibility to investigate many more types of datasets for discrete choice modeling. Indeed, representation learning methods exist for all types of inputs, including continuous signals, images, time series, and more. {Consequently, we can explore new neural network architectures (\textit{e.g.}, other variants of the residual blocks)}. Although research has already been done in this direction, \textit{e.g.}, \cite{otsuka2016deep} {who used images for choice classification}, our model is the first to propose the coexistence of these new inputs with standard discrete choice modeling variables. Furthermore, we suggest that the representation learning architecture, together with the modeled specification architecture, complement each other to reach both a better optimum when they are jointly optimized.

Secondly, we believe that other discrete choice models such as more advanced GEV models, Mixed Logit or Latent Class models can also benefit from an added data-driven term while keeping high degree of interpretability. Integrating them in a unified framework/library is a big direction for future research. Indeed, as of now, by making use of the NN structure we have been able to implement a new learnable term to the Multinomial logit, and thanks to the implementation of multiple custom loss layers in a deep learning library, we have implemented its first nested generalization, the Nested logit. Both models have benefited from a data-driven term, and we hope to bring them to more advanced models in the near future.

Moreover, the proposed architecture of our model may also help in the task of modeling the knowledge-driven term of the utility specification via feature selection. Indeed, understanding what a data-driven method has learned is an active field of research and may overall greatly benefit the field of discrete choice modeling. In our specific case, by capturing insights about the representation term, we would be able to reduce the number of parameters in $\mathcal{Q}$ incrementally, and add their interactions or non-linear functions in $\mathcal{X}$ when discovered, taking advantage of the joint optimization benefits of our model. {To this end, we may also interest ourselves to recent methods which move towards explainable A.I., notably those on network visualization and interpretation (\cite{ribeiro2016should,samek2016evaluating,shrikumar2017learning,murdoch2019definitions}) and compare with recent works on interpretable or knowledge-injected machine learning (\cite{borghesi2020improving,von2020informed,murdoch2019definitions}).}

{Yet another avenue of research concerns the structure and role of the representation term in the utility function. Indeed, for example, one could have multiple terms per utility, such that each chosen input set $\mathcal{Q}_i$, belonging to their own respective network, would create a meaningful embedding in the utility. In the same spirit, one could inspire ourselves of \cite{wang2020deep} to have an added utility specific network per alternative. On another hand, the data-driven term may simply be used to discover usually complex specifications in the utility. This has been done in \cite{han2020neural}, a direct extension of our work, which make use of neural networks to automatically discover the true taste heterogeneity function of a parameter.}

Finally, while we have shown the benefits of our data-driven term in the DCM field, the machine learning community can also benefit from our formulation to tackle small datasets. We have shown that our architecture may perform as a regularization tool for common deep learning methods when applied to small datasets. This may encourage the machine learning community to reach out for DCM practices and make use of a priori specification to help the performance of their models.

\section{Conclusions}

In this paper, we introduced a novel general and flexible theoretical framework that integrates a representation learning technique into the utility specification of a discrete choice model to automatically discover good utility specification from available data. This data-driven term may account for many forms of misspecifications and greatly improves the overall predictability of the model. Also, unlike the existing hybrid models in the literature, our framework is carefully designed to keep the interpretability of key parameters, which is critical to allow researchers and practitioners to get insights into the complex human decision-making process.

Using synthetic and real world data, we demonstrated the effectiveness of our framework by augmenting the utility specification of the Multinomial Logit, as well as the Nested Logit, with a new non-linear representation arising from a neural network, leading to new choice models referred to as the \textit{Learning Multinomial Logit} (L-MNL) and \textit{Learning Multinomial Nested Logit} (L-NL) models. Our experiments showed that our models outperformed the traditional choice models and existing hybrid models, both in terms of predictive performance and accuracy in parameter estimation. 

There is a growing interest within the transportation community to exploit multidisciplinary methods to solve the ever more challenging problems its members face.  By making our code openly available, we hope that we successfully contributed to bridge the gap between theory-driven and data-driven methods and that it will encourage researchers to combine the strengths of choice modeling and machine learning methods.

\nolinenumbers
\clearpage

\appendix

\section{Notation} \label{sec:notation}

In this section, we give a very short explanation on the main notation rule used. Variables written as plain text with or without subscripts are single elements, while their vectorized form will be in \textbf{bold} and have the relevant subscript disappear. Here are two examples: 

\begin{align}
    u_{in} &\sim (1\times1) & & & \\
    \bm{u_n} &\sim (I\times1) & or & & V_{i} &\sim (1\times1)  \\
    \bm{u_i} &\sim (1\times N) & & &\bm{V} &\sim (I \times 1) \\
    \bm{u} &\sim (I \times N) & & &
\end{align}
\color{black}

\section{Synthetic Annex Experiments } \label{sec:Synth_Annex}

\subsection{{Complex Correlation Patterns}}
\label{sec:guevara}

In this section, we make use of the \red{same} data generation process defined in \cite{guevara2015critical}. This allows us to take away one main observation: we show that a complexly correlated variable in $\mathcal{Q}$ with a variable in $\mathcal{X}$ does not negatively affect the estimated parameters. In other words, the endogeneity created by the omitted variable can be fixed when including it in the representation term without creating biasing effects. Therefore, together with other experiments of Section \ref{sec:synthetic},  we show our model fixes for omitted variable and function misspecification bias. \\

The data generation process defines two utility functions where:
\begin{equation}
 \hspace{4cm}   U_{in} = -2p_{in} + a_{in} + b_{in} +q_{in} + \varepsilon_{in} \hspace{2cm} \text{for  } i= 1,2
\end{equation}
with $p_{in}$, $a_{in}$, $b_{in}$, and $q_{in}$ being the generated variables and $\varepsilon_{in}$ is the random term. The complex correlation is between $q_{in}$ and $p_{in}$ such that: 

\begin{equation}
   \hspace{4cm}   p_{in} =  5+ q_{in} + z_{in} + 0.03wz_{in} +  \varepsilon_{pin} \hspace{2cm} \text{for  } i= 1,2
\end{equation}
where all generated variables are sampled in a uniform distribution $\mathcal{U}([-2,2])$ such that we have exactly the same data generation from the original paper.\\
As done in \cite{guevara2015critical}, we perform a Monte Carlo simulation by generating 100 repetitions of 1000 observations and by estimating the following models: 
\begin{itemize}
    \item\itemname{MNL$_{True}$:} The true logit model where $\mathcal{X} = \{\bm{a},\bm{b},\bm{q}\}$
    \item\itemname{L-MNL$_{True}$:} Complex correlated model between both sets where $\mathcal{X} = \{\bm{a},\bm{b}\}$ and  \\ \itemname{} $\mathcal{Q}=\{\bm{q}\}$. The representation term arises from a single DNN layer ($L=1$)\\ \itemname{} of  100 neurons ($H=100$). 
    \item\itemname{MNL$_{endo}$:} The endogenous logit model where $\mathcal{X} = \{\bm{a},\bm{b}\}$
\end{itemize}
All models have reach a stable optimum after 50 epochs, with standard Adam optimizer and $20\%$ dropout regularizer. Running time in this framework have no noticeable difference between models.

\begin{figure}[th]
	\begin{center}
	\includegraphics[width=0.7\textwidth,angle=0]{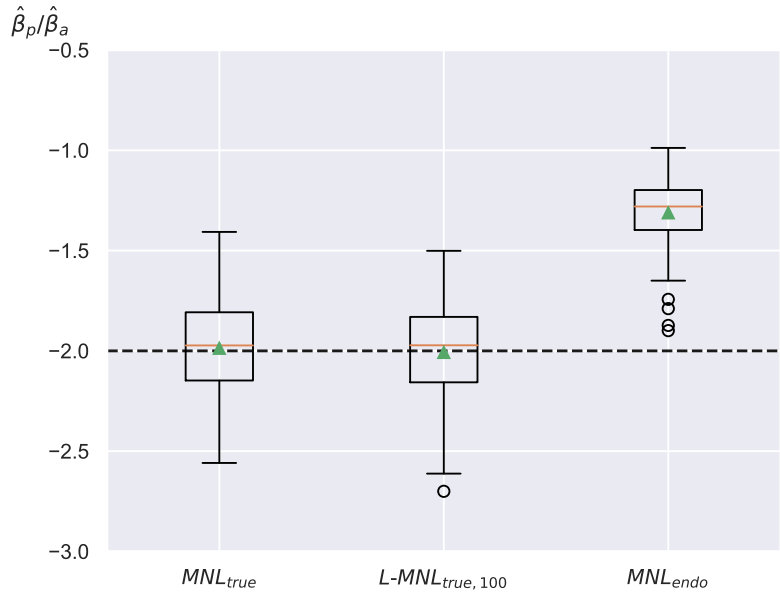}
	\end{center}
	\caption{  \label{fig:guevara} Monte Carlo experiment for complex correlation analysis, Box-Plots for 100 parameter estimates repetitions of 1000 observations each. True parameter ratio is $\beta_p/\beta_a=-2$. }
\end{figure}

The results of the experiment can be seen on Figure (\ref{fig:guevara}), where we have Box-Plots for the ratios of the estimators $\hat{\beta}_p/\hat{\beta}_a$. This value is studied since correction of endogeneity may produce a change in scale of the estimators, and we wish to remain comparable to the original paper and their multiple models. Indeed, we obtain the same distribution as in \cite{guevara2015critical} for the true model as well as the endogenous one. For the L-MNL models, we may observe that a complex correlated term in the neural network for L-MNL$_{true}$ does not affect the parameter estimates, and the overall architecture is able to estimate the true function.

\color{black}

\subsection{Impact of \red{unobserved} variables}\label{sec:unseen}
 
We take advantage of synthetic data to better analyze the impact of \red{unobserved} variables on parameter estimates. To do so, we assume that we know the following \textit{true}  utility specification for $i=1,2$: 

 \begin{equation} \label{binarylogit_systematic_new}
  V^{'}_{in} =V_{in} +\beta_u\cdot u_{in} 
 \end{equation}
 \vspace{0.15cm}
\noindent where $V_{in}$ is given by Equation (\ref{eq:binarylogit_systematic}).

We assume that variables $u_{1n},u_{2n} \sim {U}([-1,1])$ are unobserved by the modeler, \textit{i.e.}, that $\bm{u} \notin (\mathcal{X}\cup\mathcal{Q})$. We also choose $\beta_u=1$ and perform 100 experiments with this new utility specification. For the sampling scheme, we generate $1,000$ synthetic observations for the training set and $200$ more for the testing set.

Our new choice model is defined as
\begin{itemize}
\item L-MNL($100,\mathcal{X},\mathcal{Q}$), with
$\mathcal{X} = \{\bm{p},\bm{a},\bm{b}\}$, and $\mathcal{Q}=\{\bm{q},\bm{c}\}$,
\end{itemize}
and is compared with the following benchmarking model:
\begin{itemize}
\item Logit($\mathcal{X}_1$), with $\mathcal{X}_1 = \{\bm{p},\bm{a},\bm{b},\bm{q},\bm{c}\}$.
\end{itemize}
 
The Monte Carlo mean relative errors for the two models are shown in Table \ref{tab:unseen}.

\begin{table}[tb]
\centering
\caption{Impact of unseen variables on parameter estimates and log-likelihood. $\overline{e}$ is the average relative error, $s.d.$ its standard deviation and $\beta_p$, $\beta_a$ are from Equation (\ref{eq:binarylogit_systematic}).  \label{tab:unseen}}
\begin{subtable}{\linewidth}
\centering

\begin{tabular}{l|cc|cc|cc}
\textbf{Model} & \textbf{$\overline{e}_{\beta_p}$} & $s.d.(e_{\beta_p})$ & \textbf{$\overline{e}_{\beta_a}$} & $s.d.(e_{\beta_a})$ & \textbf{$\overline{e}_{\beta_p/\beta_a}$} & $s.d.(e_{\beta_p/\beta_a})$ \\
\hline
L-MNL($\red{25},\mathcal{X},\mathcal{Q}$) & $\bm{11.2}$& \pmtabbf{$\bm{7.4}$}& $\bm{21.1}$  & \pmtabbf{$\bm{13.6}$} & $\bm{16.2}$ & \pmtabendbf{$\bm{12.5}$} \\[1pt]
\hline
Logit($\mathcal{X}_1$) & 34.2 & \pmtab{6.3} & 38.7 & \pmtab{16.2}&  20.9& \pmtabend{15.5} \\[1pt]
\hline
\end{tabular}
\end{subtable}
\par\medskip
\begin{subtable}{\linewidth}
\centering
\hspace{-3.15cm}

\begin{tabular}{l|cc|cc}
 & $\overline{LL}_{train}$ & $s.d.(LL)$ &  $\overline{LL}_{test}$ & $s.d.(LL)$ \\
\hline
L-MNL($\red{25},\mathcal{X},\mathcal{Q}$) & $\bm{-460}$ & \pmtabbf{$\bm{16}$} & $\bm{-101}$ & \pmtabendbf{$\bm{7}$}\\[1pt]
\hline
Logit($\mathcal{X}_1$) & -620& \pmtab{11} & -124 & \pmtabend{5}\\[1pt]
\hline
\end{tabular}
\end{subtable}
\end{table}

We see that, compared to the results depicted in Table \ref{tab:montecarlo_errors} where all variables in the \textit{true} utility specification had been included in the models, the unseen variables lead, for both models, to a decrease in the fit and an increase in relative errors in parameter estimates. Nevertheless, { one important observation is that our L-MNL retrieves consistent ratio estimates, similarly to the MNL model, while having better global performances. The unseen variables do not provoke unexpected behavior to our new model.  }

\subsection{Optimization Strategy} \label{sec:joint_optimization}

As depicted in Figure \ref{fig:hyb}, our L-MNL architecture contains two parts: the top of the figure represents the \textit{a priori} defined component in the utility specification, while the bottom depicts the learning component. The number of parameters to be estimated in each part can vary significantly. While a limited number of parameters are generally included in linear specification, a neural network can easily contain thousands of parameters. Given this architecture, one could be tempted to estimate the model sequentially. To show the importance of jointly estimating the parameters, we consider the three following strategies for estimating our L-MNL($100,\mathcal{X},\mathcal{Q}$):

\begin{itemize}
\item[1)] First optimizing the standard discrete choice component of the specification (\textit{i.e.}, the $\bm{\beta}$ parameters) and then learning the representation term after fixing the previously found linear-in-parameter estimates.
\item[2)] First optimizing the representation term (\textit{i.e.}, the $\bm{w}$ weights) and then learning the modeled specification after fixing the previously estimated weights.
\item[3)] Optimizing jointly both components. 
\end{itemize}

Results are shown in Table \ref{tab:optimization} for $\beta_p=-2$, $\beta_a=1$,$\beta_b=0.5$ and $\beta_{qc}=1$. We see that the joint optimization allows for the best minima in both likelihood and parameter estimation. Starting with the modeled specification gives the same parameters as if it were an MNL, but ends up with a better likelihood as the NN component then only increases predictability. The second strategy, on the other hand, reaches a sub-optimal minima when learning the representation. It is only with joint optimization that all important explanatory variables are expressed. The two components complete each other to achieve the best prediction performance with the correct parameter values. 

\begin{table}[hbt]
\centering
\caption{Values of parameter estimates and likelihoods based on optimization strategy for $\beta_p=-2$ and $\beta_a= 1$. Best results are obtained with joint optimization.  \label{tab:optimization}}

\begin{tabular}{l|cc|cc}
 Strategy & $\hat{\beta}_p$ & $\hat{\beta}_a$   & $\mathcal{L}(\hat{\beta})$ & $\mathcal{L}_{test}(\hat{\beta})$ \\[3pt]
\hline
& & & \\[-8pt]
(1)  $\bm{\hat{\beta}}$ then 	$\bm{w}$ & -$1.41$ & $0.75$ & -4927 & -942\\
(2) $\bm{w}$  then $\bm{\hat{\beta}}$	&-$1.59$ & $0.82$& -3924 & -758 \\
(3)  $\bm{\hat{\beta}}$ and 	$\bm{w}$  & -$\bm{1.95}$& $\bm{1.0}$ & -$\bm{3678}$  & -$\bm{721}$\\[3pt]
\hline
\end{tabular}
\end{table}

\section{Semi-Synthetic Data} \label{sec:semi_synthetic}

In this section, we challenge our L-MNL model by analyzing its effectiveness outside of fully synthetic data and in the presence of strong non-linearities and noisy data. To generate our semi-synthetic data, we follow the same procedure as in section $\ref{sec:synthetic}$, but instead of using normally distributed explanatory variables, we randomly select them from a real world dataset, the Swissmetro dataset (\cite{bierlaire2001acceptance}). 

We define the following utility specifications:

\begin{alignat}{5} \nonumber 
&V_{Train} &=&  \hspace{5pt} l_{Train} &&  \hspace{5pt} + 1 \cdot \textit{{\scriptsize{$DEST$}}}^3\cdot \textit{{\scriptsize{$AGE$}}} && - 1 \cdot \textit{{\scriptsize{$AGE$}}}^{0.5}\cdot \textit{{\scriptsize{$ORIGIN$}}} \\  
&V_{SM} &=&  \hspace{5pt} l_{SM} &&  \hspace{5pt} + 1\cdot \textit{{\scriptsize{$DEST$}}}\cdot \textit{{\scriptsize{$AGE$}}}  &&+ 3\cdot \textit{{\scriptsize{$INCOME$}}}^5 \cdot \textit{{\scriptsize{$PURPOSE$}}}^2 \nonumber \\
&V_{Car} &=& \hspace{5pt} l_{Car} && \hspace{5pt} +5\cdot \textit{{\scriptsize{$AGE$}}}\cdot\textit{{\scriptsize{$INCOME$}}}^5 && +2 \cdot \textit{{\scriptsize{$ORIGIN$}}}^2\cdot \textit{{\scriptsize{$INCOME$}}}^5
\label{eq:semi_synth_complex}
\end{alignat}
with 
\begin{equation} \label{eq:semi_synth_linear}
l_{i} = -1\cdot \textit{{\footnotesize{$TT_{i}$}}} -2\cdot \textit{{\footnotesize{$TC_{i}$}}} \hspace{30pt} \forall i\in\mathcal{C},
\end{equation}
The coefficients are chosen in order to have a balanced dataset, and the interacting variables are the categorical features of Swissmetro data (\cite{bierlaire2001acceptance}) and are described in Table \ref{datasemisyntethic}. Note that we have chosen to use power series to complexify the non-linear terms. 

The models under study are: 
\begin{itemize}
\item Logit($\mathcal{X}_{a}$) with linear-in parameters utility specification based on the following features: \begin{itemize}
	\item $\mathcal{X}_{Train,a}=\{${\footnotesize{1, $TT_{Train}, TC_{Train}, AGE, DEST, ORIGIN$}}$\}$
    \item$\mathcal{X}_{SM,a} = \{${\footnotesize{1, $TT_{SM}, TC_{SM}, AGE, DEST, INCOME, PURPOSE$}}$\}$
    \item$\mathcal{X}_{Car,a} = \{${\footnotesize{1, $TT_{Car}, TC_{Car}, AGE, ORIGIN, INCOME $}}$\}$
    \end{itemize}
\item Logit($\mathcal{X}_b$) with only travel time and cost for each utility, \textit{i.e.}: 
\begin{itemize}
	\item $\mathcal{X}_{ib} = \{${\footnotesize{$1, TT_i, TC_i$}}$\}$ for all $i\in \mathcal{C}$.
  \end{itemize}
\item L-MNL($\mathcal{X}$, $\mathcal{Q}$) with 
\begin{itemize}
	\item $\mathcal{X}_i = \{${\footnotesize{$TT_i, TC_i$}}$\}$, 
\item $\mathcal{Q}=\{${\footnotesize{$AGE,DEST,ORIGIN,INCOME,PURPOSE$}}$\}$
\end{itemize}
   \end{itemize}

The results can be seen in Table \ref{tab:semi_synth_lmnl_wins}. We see that standard MNL models are unable to retrieve the correct parameter estimates for utility specifications with important non-linearities. Both MNL models exhibit large relative errors (about $40\%$ for the Logit($\mathcal{X}_b$) and at least $25\%$ for the Logit($\mathcal{X}_a$)). The relative errors are also large for the ratio of parameters, which would lead to wrong postestimation indicator, the VOT in this case. Unlike the Logit models, our L-MNL model recovers the true estimates in both parameters and ratio, while achieving a much better fit. We therefore conclude that the representation term was able to learn the complex non-linearities and that ignoring these non-linearities lead to models that greatly suffer from underfit. 

\begin{table}[tb]
\centering
\caption{Values of parameter estimates and likelihoods for different models based on Equation (\ref{eq:semi_synth_complex}). Ground truth is $\beta_{TT}=-1$ and $\beta_{TC}= -2$. Only L-MNL is able to estimate correctly the parameters. \label{tab:semi_synth_lmnl_wins}}
\begin{tabular}{l|ccc|cc}
 Models & $\hat{\beta}_{TT}$ & $\hat{\beta}_{TC}$  & $\hat{\beta}_{TC}/\hat{\beta}_{TT}$ & $LL_{train}$ & $LL_{test}$  \\[3pt]
\hline
 & & & \\[-8pt]
Logit($\mathcal{X}_a$) & -0.65& -1.63&  2.50 & -5412 & -1354\\
Logit($\mathcal{X}_b$) &  -0.25 & -0.70 & 2.81 & -7722& -1925\\
L-MNL($\mathcal{X}, \mathcal{Q}$) & $\bm{-1.01}$& $\bm{-1.99}$ & $\bm{1.96}$ &  $\bm{-2516}$ & $\bm{-809}$\\
\hline
\end{tabular}
\end{table}

\section{Swissmetro supplementary experiment}
\subsection{Feature impact and sensitivity analysis} \label{sec:sensitivity}
\begin{figure}[tb]
\centering
\hspace{-1.5cm}\begin{subfigure}[t]{0.5\textwidth}
	\caption{}
	\includegraphics[width=1\textwidth,angle=0]{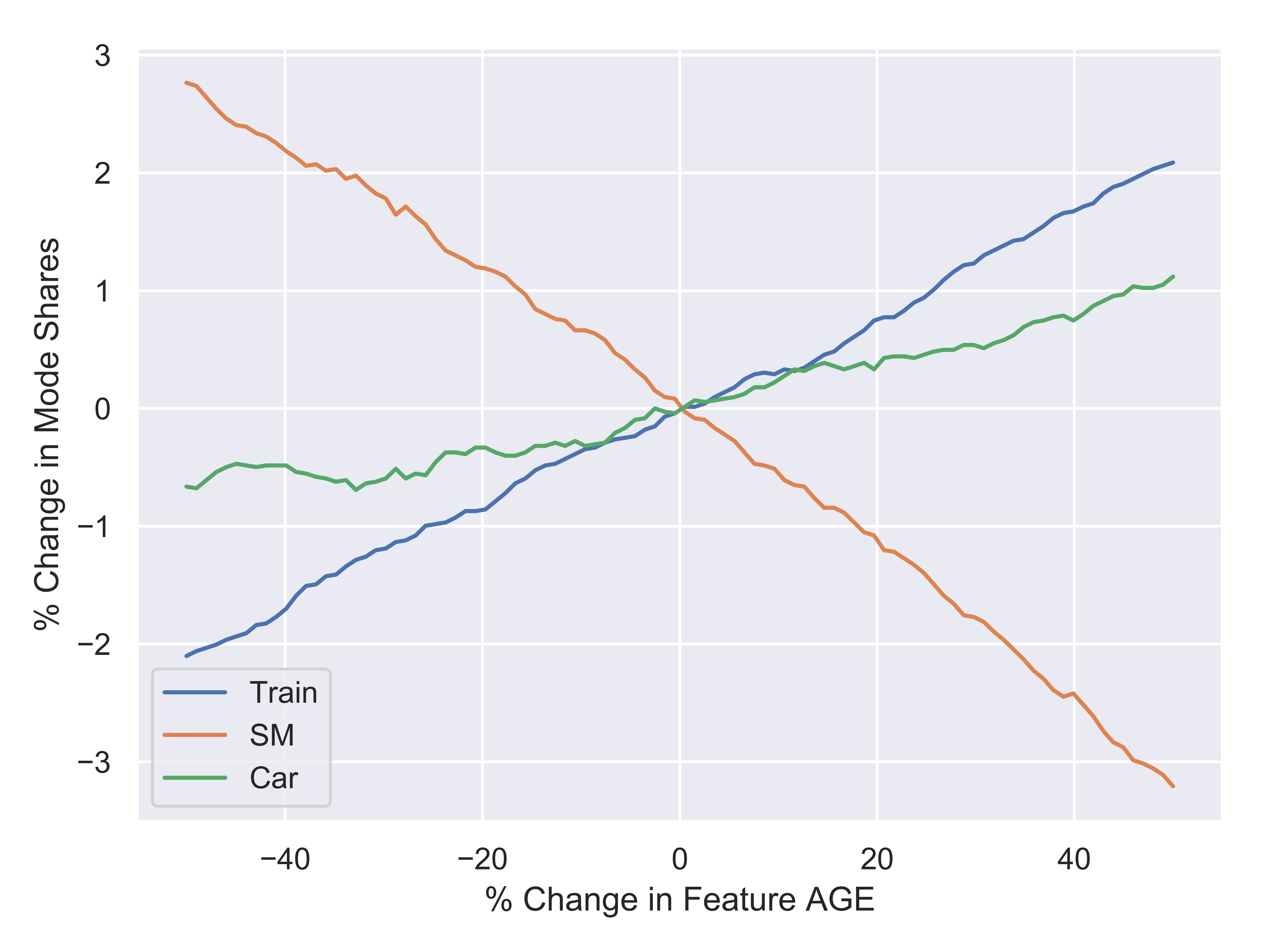}
\end{subfigure}
\begin{subfigure}[t]{0.5\textwidth}
	\caption{}
	\includegraphics[width=1.\textwidth,angle=0]{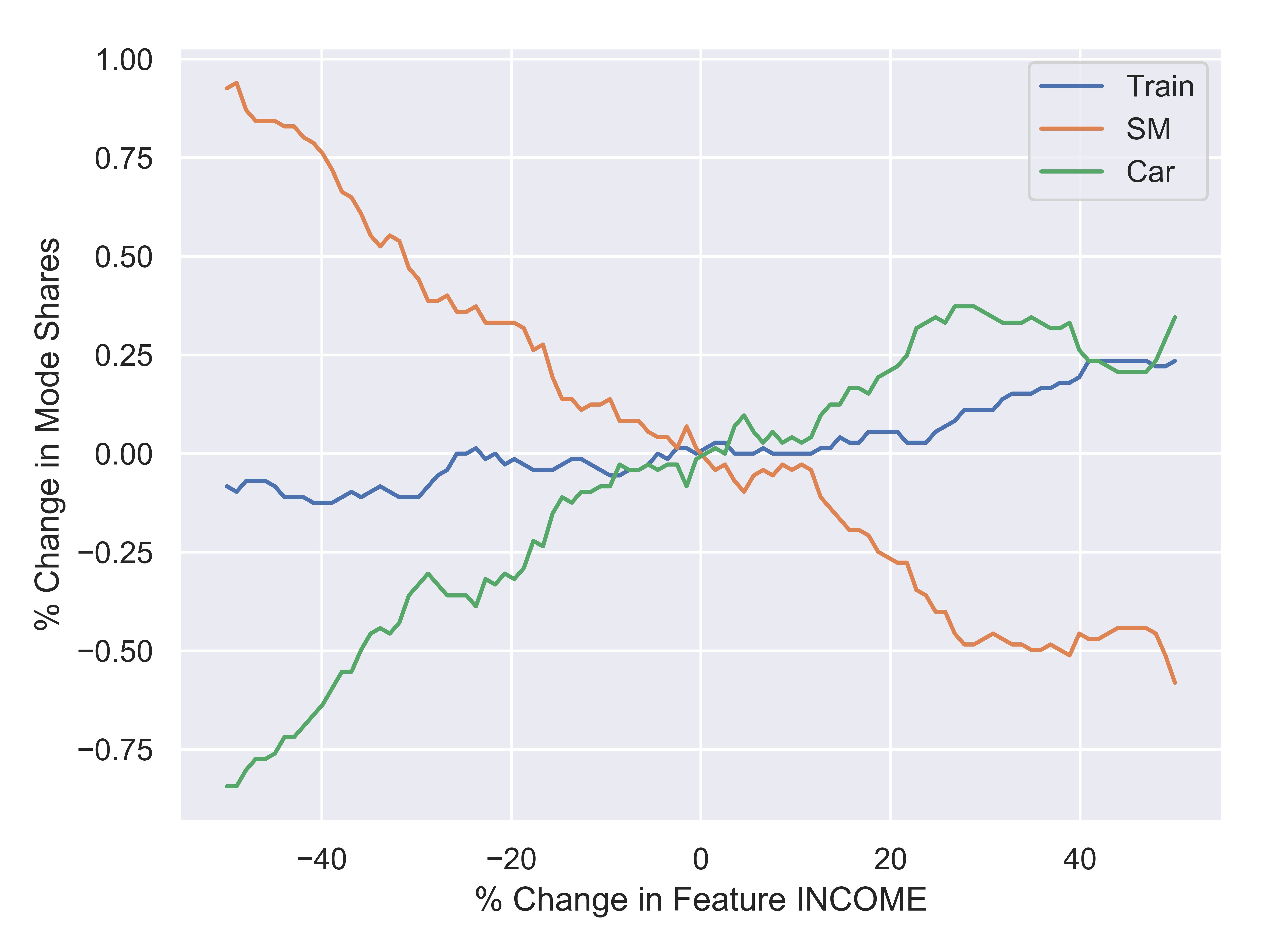}
\end{subfigure}
\caption{Percentage change in total mode shares against percentage change in L-MNL features belonging to $\mathcal{Q}_2$. These general trends appear by fixing all other values as constant.} \label{fig:sensitivity}
\end{figure}

As previously done by \cite{bentz2000neural}, we finish our experiments by investigating what the neural network component has learned through the study of a sensitivity analysis. To do so, we vary the value of a feature in $\mathcal{Q}$ while keeping the others variables constant, and we analyze its impact on the utilities and market shares of the alternatives. We do the analysis on the L-MNL($100,\mathcal{X}_2,\mathcal{Q}_2$) for which only $cost$, $time$, and $frequency$ are included in the linear specification, while the other $14$ variables are given to the neural network. Figure \ref{fig:sensitivity} presents a sensitivity analysis for two variables in $\mathcal{Q}_2$: $AGE$ and $INCOME$\footnote{These two variables were chosen for elasticity study as they are the only integer variables which represent a discrete scale of intensity. }. We observe that the $AGE$ variable has almost linear relations to the utilities, which has also been seen in \cite{bierlaire2001acceptance}'s benchmark. Changing $INCOME$ however, seems to present non-linearities and an overall weaker impact on the change in mode share. We recognize that this is only the average behavior of the feature in the population when keeping all other variables constant. Further investigation of non-linear interactions can be done by separating our sensitivity analysis based on the values of the other variables as seen in \cite{bentz2000neural}.

To have some insights on the impact of each feature in $\mathcal{Q}$ on the utility function, we get inspiration from saliency maps \cite{simonyan2013deep}. A saliency map is obtained through back-propagation of an observation's prediction score, as opposed to its output loss which is performed during training. We then read the results at the nodes of the input layer. The retrieved values are considered to be the gradient estimation of a prediction with respect to a given input. For our case, we do this by changing the loss function of our pre-trained model: 
\begin{equation} \label{eq:feature_loss}
loss(\bm{V}_n,\bm{y}_n ) = \sum\limits_{i\in\mathcal{C}_n} y_{in}\cdot V_{in},
\end{equation}
 where $y_{in}$ is 1 when individual $n$ is predicted to choose alternative $i$ and $V_{in}$ is the output for utility $i$ as seen in Equation (\ref{eq:systematicutility}). 
 
\begin{figure}[tb]
	\begin{center}
	\includegraphics[width=0.7\textwidth,angle=0]{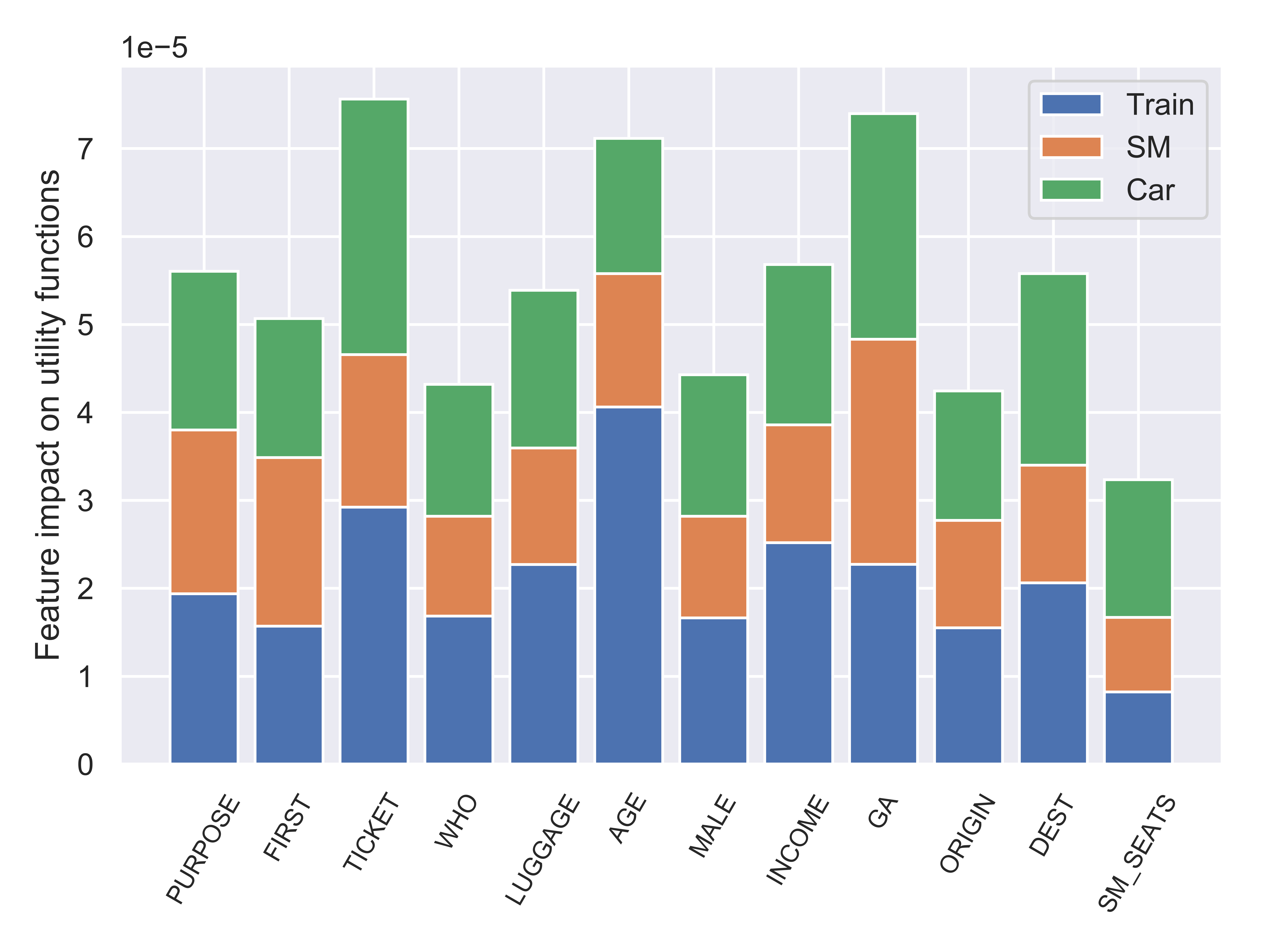}
	\end{center}
	\caption{  \label{fig:feature_impact} Mean feature contributions to each utility obtained with loss of Equation (\ref{eq:feature_loss}) and gradient evaluation on the input layer.    }
\end{figure}

As opposed to an image, which is a 2D set of pixels, the position of our input has always the same meaning for each individual. In other words, the gradient read on the first position will always be $PURPOSE$ and the last always $SM\_SEATS$. This allows us to measure the average impact of a feature on each class. Indeed, when summing over all individuals the absolute gradients at the input layer for each feature, we get the Figure \ref{fig:feature_impact}. The sum has been separated for each utility and normalized by the count of the chosen alternative of all individuals.

As we can see, all variables are being used by the NN. Some, such as $GA$, $AGE$ or $LUGGAGE$ seem to have an overall bigger impact than $SM\_SEATS$. This supports the conclusion that the MNL benchmark of \cite{bierlaire2001acceptance} misses potential useful information by ignoring many variables. 

\clearpage
{
\section{Learning Nested-Logit: network-loss architecture} \label{sec:nested_loss}
\begin{figure}[H]
	\begin{center}
	\includegraphics[width=0.7\textwidth,angle=0]{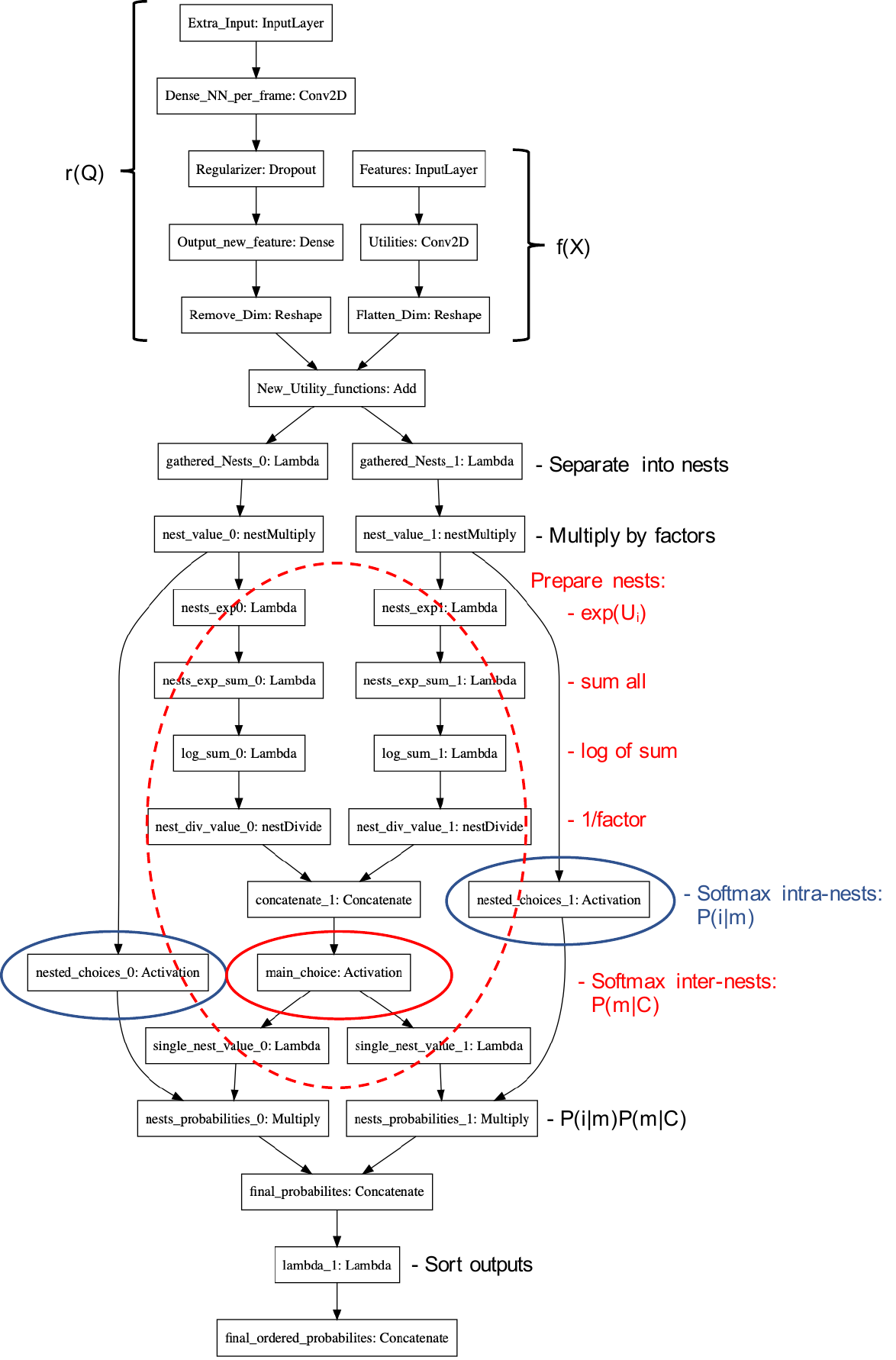}
	\end{center}
	\caption{  \label{fig:nested_loss} For 2 nests. As the nested loss needs trainable weights, this was solved using custom layers. "nestMultiply" and "nestDivide" share the same trainable parameter $\mu$ and has constraint $\mu\geq1$      }
\end{figure} 
}
\clearpage
\section{\red{Extra Models details for Swissmetro}} \label{sec:SM_extra_models}
 \begin{table}[h]
 \centering
  \caption{Comparison of parameters estimates for different models with utility specification of \cite{bierlaire2001acceptance} }
  \begin{subtable}{.45\linewidth}
\caption{Parameter estimates from CNL($\mathcal{X}_1$) model}
  \begin{adjustbox}{width=\linewidth}
\begin{tabular}{lllll}
Parameters &  Value &  Std errors &  t-test &  p-value \\ \hline
$\alpha_{existing}$ &  0.371 &   0.0229 &    16.2 &  0.0 \\
$ASC_{Car}$  & -0.015 &    0.045 &  -0.333 &    0.739  \\
$ASC_{Train}$  & 0.0245 &   0.0793 &   0.309 & 0.757 \\
 $\beta_{age}$  &  0.108 &   0.0197 &    5.48 & 4.18e-08 \\
 $\beta_{cost}$ &  -0.58 &    0.037 &   -15.7 &  0.0\\
 $\beta_{freq}$ &  -0.38 &   0.0591 &   -6.43 & 1.27e-10  \\
 $\beta_{GA}$   &   1.28 &    0.123 &    10.4 &  0.0\\
 $\beta_{luggage}$  & -0.136 &   0.0395 &   -3.44 & 0.000591 \\
 $\beta_{seats}$    & 0.0977 &   0.0738 &    1.32 &    0.185 \\
 $\beta_{time}$  & -0.982 &   0.0535 &   -18.3 &  0.0 \\
$\mu_{existing}$    &    1.8 &    0.124 &    14.5 &  0.0 \\
$\mu_{public}$  &   4.57 &    0.566 &    8.07 & 6.66e-16\\ \hline
\multicolumn{2}{l}{Number of observations }&  7,234& & \\
\multicolumn{2}{l}{$\mathcal{L}(\hat{\beta})$ = -5711 }& & \multicolumn{2}{l}{$\mathcal{L}_{test}(\hat{\beta}) = -1415$ }
\end{tabular}
\end{adjustbox}
\end{subtable}\hspace{0.8cm}
\begin{subtable}{.45\linewidth}
\caption{Parameter estimates from TPM($\mathcal{X}_1$) model}
  \begin{adjustbox}{width=\linewidth}
\begin{tabular}{lllll}
Parameters &  Value &  Std errors &  t-test &  p-value \\ \hline
$ASC_{Car}$     &  -0.675 &    0.146 &   -4.61 & 3.94e-06 \\
 $ASC_{Train}$ &   -2.52 &    0.245 &   -10.3 &      0.0 \\
 $\beta_{age}$    &   0.444 &    0.066 &    6.73 & 1.75e-11 \\
 $\beta_{cost}$     &   -1.56 &    0.103 &   -15.1 &      0.0  \\
 $\beta_{freq}$      &  -0.981 &     0.14 &   -6.98 & 2.86e-12 \\
 $\beta_{GA}$        &    1.03 &    0.467 &     2.2 &   0.0278  \\
 $\beta_{luggage}$  & -0.0687 &    0.161 &  -0.427 &     0.67 \\
 $\beta_{seats}$     &  0.0373 &    0.127 &   0.293 &    0.769  \\
 $\beta_{time}$      &   -2.17 &   0.0913 &   -23.7 &      0.0 \\
$\sigma_{Car}$   &    2.81 &    0.159 &    17.7 &      0.0  \\
$\sigma_{Train}$ &   -1.41 &   0.0921 &   -15.3 &      0.0 \\
\hline
\multicolumn{2}{l}{Number of observations }&  7,234& & \\
\multicolumn{2}{l}{$\mathcal{L}(\hat{\beta})$ = -4752 }& & \multicolumn{2}{l}{$\mathcal{L}_{test}(\hat{\beta}) = -1350$ }
\end{tabular}
\end{adjustbox}
\end{subtable}
\end{table}
\clearpage
\section{Correlation coefficients among variables in Swissmetro dataset}\label{sec:correlation_SM}
  \begin{center}
  \scriptsize
  \hvFloat[%
nonFloat=true,
objectAngle=90,%
capPos=r,%
capAngle=90,
capWidth=0.5]{table}{%
\begin{tabular}{|c|ccccccccccccccc|}
  \toprule
\hline
& Purpose & First & 	Ticket&	Who &	Origin &	Dest &	Male &	Income &	GA &	Luggage &	Age	& Seats &	TT	 & TC	& Freq \\
 \hline
Purpose	 &  1.00 & 	-0.08 & 	-0.18 & 	-0.19 & 	-0.02 & 	0.05 & 	-0.01 & 	-0.08 & 	-0.13 & 	-0.01 & 	0.11 & 	-0.07 & 	0.16 & 	-0.13 & 	-0.06 \\
First	& 	& 1.00 & 	-0.11 & 	0.21 & 	-0.08 & 	0.00 & 	0.21 & 	0.24 & 	-0.07 & 	-0.10 & 	0.16 & 	-0.06 & 	0.05 & 	0.03 & 	-0.05 \\
Ticket		& & &	1.00 & 	0.00 & 	0.04	 & 0.15 & 	-0.11 & 	0.00 & 	0.55 & 	0.20 & 	-0.11& 	0.26	& -0.12& 	0.50	& 0.15\\
Who			& & & &	1.00 &	-0.06	&-0.04&	0.16	&0.31	&0.05&	0.00&	-0.07&	0.00&	-0.01&	0.07&	-0.02\\
Origin			& & & &&		1.00&	-0.12	&-0.10&	-0.02&	-0.09&	-0.06&	0.01&	-0.02&	0.12&	-0.08&	0.01\\
Dest				& & & && &		1.00&	-0.05	&0.01&	0.08&	0.12&	0.02&	0.06&	0.16&	0.07&	0.06\\
Male 					& & & && & &		1.00&	0.11&	-0.04&	-0.16	&0.11&	-0.15&	0.04&	0.00&	-0.08\\
Income					& & && & & & &			1.00&	0.00	&0.05&	0.10&	0.00&	0.04&	0.04&	0.01\\
GA							& & & & && & & &		1.00	&0.23	&-0.06	&0.26	&-0.14&	0.90&	0.23\\
Luggage					& & & & && & & & & 					1.00&	-0.05&	0.18&	0.03&	0.21&	0.10\\
Age							& & & & & && & & & &				1.00	&-0.06&	0.13&	-0.05&	0.00\\
Seats								& & & && & & & & & & &			1.00	&-0.14	&0.25&	0.17\\
TT							& & & & & & && & & & & &						1.00 &	-0.15 &	-0.09\\
TC					& & & & & & & & & && & & &									1.00 &	0.20\\
Freq				& & & & & & & & & & && & & &											1.00\\
\hline
  \bottomrule
\end{tabular}
}{Correlation coefficients among variables included in Swissmetro for our training dataset}{tab:coeffcorrelation}
  \end{center}
\normalsize
\clearpage

\section{\red{Correlation coefficients among variables in Optima dataset}}\label{sec:correlation_optima}
\vspace*{3cm}
  \begin{center}
  \scriptsize
  \hvFloat[%
nonFloat=true,
objectAngle=90,%
capPos=r,%
capAngle=90,
capWidth=0.8]{table}{%
\begin{tabular}{|c|cccccccc|}
\toprule
\hline
 &  age & HouseType & Gender  & Income & Educ. & SocioCat & House & Famil.  \\
 \hline
TimePT &  -0.07 &   0.04  & -0.03  &   0.07  &   0.09  & -0.01 &   0.09  &   0.08  \\
TimeCar & -0.00 &  0.00  & -0.06 &   0.04  &   0.07  & -0.03 &   0.07  &   0.08   \\
CostPT &  -0.02 &  0.02  & -0.06  &   0.04  &   0.06  & -0.03 &   0.08  &   0.06   \\
distance &  0.01  -0.05 & -0.01 & -0.25 &   0.03  &   0.06  & -0.04 &   0.06  &   0.07   \\
CostCarCHF &  0.01 &  -0.01 & -0.05  &  0.03 &  0.07 & -0.04 &  0.06 &  0.07  \\
TripPurpose &  0.23 & -0.02 &  0.08 & -0.11 &  0.03 &  0.06 & -0.04 & -0.09  \\
NbBicy & -0.38 &  -0.18 &  0.03 &  0.24 &  0.05 & -0.01 & -0.09 &  0.35  \\
NbCar &  -0.16 &  -0.17 & -0.04 &    0.24 & -0.04 & -0.04 & -0.08 &  0.20  \\
NbChild & -0.41 &  -0.07 &  0.07 &   0.13 &  0.09 &  0.01 & -0.02 &  0.17  \\
GenAbST & 0.06 & 0.06 &  0.04 &  -0.00 & -0.06 &  0.02 & -0.01 & -0.14 \\ \hline
\bottomrule
\end{tabular}
}{Correlation coefficients among variables between sets $\mathcal{X}$ and $\mathcal{Q}$ included in Optima for our training dataset}{tab:coeffcorrelation_Optima}
 \end{center}
\normalsize
\clearpage
 
\bibliographystyle{abbrvnat}
\bibliography{main}

\begin{thebibliography}{85}
\providecommand{\natexlab}[1]{#1}
\providecommand{\url}[1]{\texttt{#1}}
\expandafter\ifx\csname urlstyle\endcsname\relax
  \providecommand{\doi}[1]{doi: #1}\else
  \providecommand{\doi}{doi: \begingroup \urlstyle{rm}\Url}\fi

\bibitem[Abe(1999)]{abe1999generalized}
M.~Abe.
\newblock A generalized additive model for discrete-choice data.
\newblock \emph{Journal of Business \& Economic Statistics}, 17\penalty0
  (3):\penalty0 271--284, 1999.

\bibitem[Ackley et~al.(1985)Ackley, Hinton, and Sejnowski]{ackley1985}
D.~H. Ackley, G.~E. Hinton, and T.~J. Sejnowski.
\newblock A learning algorithm for boltzmann machines.
\newblock \emph{Cognitive science}, 9\penalty0 (1):\penalty0 147--169, 1985.

\bibitem[Agrawal and Schorling(1996)]{agrawal1996}
D.~Agrawal and C.~Schorling.
\newblock Market share forecasting: An empirical comparison of artificial
  neural networks and multinomial logit model.
\newblock \emph{Journal of Retailing}, 72\penalty0 (4):\penalty0 383--407,
  1996.

\bibitem[Bentz and Merunka(2000)]{bentz2000neural}
Y.~Bentz and D.~Merunka.
\newblock Neural networks and the multinomial logit for brand choice modelling:
  a hybrid approach.
\newblock \emph{Journal of Forecasting}, 19\penalty0 (3):\penalty0 177--200,
  2000.

\bibitem[Berger et~al.(1999)Berger, Liseo, and Wolpert]{brunero}
J.~O. Berger, B.~Liseo, and R.~L. Wolpert.
\newblock Integrated likelihood methods for eliminating nuisance parameters.
\newblock \emph{Statistical Science}, 14\penalty0 (1):\penalty0 1--22, 1999.
\newblock ISSN 08834237.

\bibitem[Bierlaire(2003)]{bierlaire2003biogeme}
M.~Bierlaire.
\newblock Biogeme: a free package for the estimation of discrete choice models.
\newblock In \emph{Swiss Transport Research Conference}, number CONF, 2003.

\bibitem[Bierlaire et~al.(2001)Bierlaire, Axhausen, and
  Abay]{bierlaire2001acceptance}
M.~Bierlaire, K.~Axhausen, and G.~Abay.
\newblock The acceptance of modal innovation: The case of swissmetro.
\newblock \penalty0 (TRANSP-OR-CONF-2006-055), 2001.

\bibitem[Bierlaire et~al.(2011)Bierlaire, Curchod, Danalet, Doyen, Faure,
  Glerum, Kaufmann, Tabaka, and Schuler]{bierlaire2011projet}
M.~Bierlaire, A.~Curchod, A.~Danalet, E.~Doyen, P.~Faure, A.~Glerum,
  V.~Kaufmann, K.~Tabaka, and M.~Schuler.
\newblock Projet de recherche sur la mobilit{\'e} combin{\'e}e, rapport
  d{\'e}finitif de l'enqu{\^e}te de pr{\'e}f{\'e}rences r{\'e}v{\'e}l{\'e}es.
\newblock Technical report, 2011.

\bibitem[Bishop(1995)]{bishop1995neural}
C.~M. Bishop.
\newblock \emph{Neural networks for pattern recognition}.
\newblock Oxford university press, 1995.

\bibitem[Borghesi et~al.(2020)Borghesi, Baldo, and
  Milano]{borghesi2020improving}
A.~Borghesi, F.~Baldo, and M.~Milano.
\newblock Improving deep learning models via constraint-based domain knowledge:
  a brief survey.
\newblock \emph{arXiv preprint arXiv:2005.10691}, 2020.

\bibitem[Brathwaite et~al.(2017)Brathwaite, Vij, and
  Walker]{brathwaite2017machine}
T.~Brathwaite, A.~Vij, and J.~L. Walker.
\newblock Machine learning meets microeconomics: The case of decision trees and
  discrete choice.
\newblock \emph{Working paper arXiv preprint arXiv:1711.04826}, 2017.

\bibitem[Breiman(1996)]{breiman1996}
L.~Breiman.
\newblock Bagging predictors.
\newblock \emph{Machine learning}, 24\penalty0 (2):\penalty0 123--140, 1996.

\bibitem[Breiman(2001)]{breiman2001}
L.~Breiman.
\newblock Random forests.
\newblock \emph{Machine learning}, 45\penalty0 (1):\penalty0 5--32, 2001.

\bibitem[Cantarella and de~Luca(2005)]{cantarella2005multilayer}
G.~E. Cantarella and S.~de~Luca.
\newblock Multilayer feedforward networks for transportation mode choice
  analysis: An analysis and a comparison with random utility models.
\newblock \emph{Transportation Research Part C: Emerging Technologies},
  13\penalty0 (2):\penalty0 121--155, 2005.

\bibitem[Chang et~al.(2019)Chang, Wu, Liu, Yan, Sun, and Qu]{chang2019travel}
X.~Chang, J.~Wu, H.~Liu, X.~Yan, H.~Sun, and Y.~Qu.
\newblock Travel mode choice: a data fusion model using machine learning
  methods and evidence from travel diary survey data.
\newblock \emph{Transportmetrica A: Transport Science}, 15\penalty0
  (2):\penalty0 1587--1612, 2019.

\bibitem[Chollet et~al.(2015)]{chollet2015keras}
F.~Chollet et~al.
\newblock Keras.
\newblock \url{https://github.com/fchollet/keras}, 2015.

\bibitem[Cortes and Vapnik(1995)]{cortes1995}
C.~Cortes and V.~Vapnik.
\newblock Support-vector networks.
\newblock \emph{Machine learning}, 20\penalty0 (3):\penalty0 273--297, 1995.

\bibitem[Cramer(2007)]{cramer2007robustness}
J.~S. Cramer.
\newblock Robustness of logit analysis: Unobserved heterogeneity and
  mis-specified disturbances.
\newblock \emph{Oxford Bulletin of Economics and Statistics}, 69\penalty0
  (4):\penalty0 545--555, 2007.

\bibitem[Dougherty(1995)]{dougherty1995}
M.~Dougherty.
\newblock A review of neural networks applied to transport.
\newblock \emph{Transportation Research Part C: Emerging Technologies},
  3\penalty0 (4):\penalty0 247--260, 1995.

\bibitem[Draxler et~al.(2018)Draxler, Veschgini, Salmhofer, and
  Hamprecht]{draxler2018essentially}
F.~Draxler, K.~Veschgini, M.~Salmhofer, and F.~A. Hamprecht.
\newblock Essentially no barriers in neural network energy landscape.
\newblock \emph{arXiv preprint arXiv:1803.00885}, 2018.

\bibitem[Faghri and Hua(1992)]{faghri1992}
A.~Faghri and J.~Hua.
\newblock Evaluation of artificial neural network applications in
  transportation engineering.
\newblock \emph{Transportation Research Record}, 1358:\penalty0 71, 1992.

\bibitem[Fern{\'a}ndez-Antol{\'\i}n et~al.(2016)Fern{\'a}ndez-Antol{\'\i}n,
  Guevara, De~Lapparent, and Bierlaire]{fernandez2016correcting}
A.~Fern{\'a}ndez-Antol{\'\i}n, C.~A. Guevara, M.~De~Lapparent, and
  M.~Bierlaire.
\newblock Correcting for endogeneity due to omitted attitudes: Empirical
  assessment of a modified mis method using rp mode choice data.
\newblock \emph{Journal of choice modelling}, 20:\penalty0 1--15, 2016.

\bibitem[Friedman(2001)]{friedman2001}
J.~H. Friedman.
\newblock Greedy function approximation: a gradient boosting machine.
\newblock \emph{Annals of statistics}, pages 1189--1232, 2001.

\bibitem[Garipov et~al.(2018)Garipov, Izmailov, Podoprikhin, Vetrov, and
  Wilson]{garipov2018loss}
T.~Garipov, P.~Izmailov, D.~Podoprikhin, D.~P. Vetrov, and A.~G. Wilson.
\newblock Loss surfaces, mode connectivity, and fast ensembling of dnns.
\newblock In \emph{Advances in Neural Information Processing Systems}, pages
  8789--8798, 2018.

\bibitem[Glerum et~al.(2014)Glerum, Atasoy, and Bierlaire]{glerum2014using}
A.~Glerum, B.~Atasoy, and M.~Bierlaire.
\newblock Using semi-open questions to integrate perceptions in choice models.
\newblock \emph{Journal of choice modelling}, 10:\penalty0 11--33, 2014.

\bibitem[Golshani et~al.(2018)Golshani, Shabanpour, Mahmoudifard, Derrible, and
  Mohammadian]{golshani2018modeling}
N.~Golshani, R.~Shabanpour, S.~M. Mahmoudifard, S.~Derrible, and
  A.~Mohammadian.
\newblock Modeling travel mode and timing decisions: Comparison of artificial
  neural networks and copula-based joint model.
\newblock \emph{Travel Behaviour and Society}, 10:\penalty0 21--32, 2018.

\bibitem[Guevara(2015)]{guevara2015critical}
C.~A. Guevara.
\newblock Critical assessment of five methods to correct for endogeneity in
  discrete-choice models.
\newblock \emph{Transportation Research Part A: Policy and Practice},
  82:\penalty0 240--254, 2015.

\bibitem[Hagenauer and Helbich(2017)]{hagenauer2017}
J.~Hagenauer and M.~Helbich.
\newblock A comparative study of machine learning classifiers for modeling
  travel mode choice.
\newblock \emph{Expert Systems with Applications}, 78:\penalty0 273--282, 2017.

\bibitem[Han et~al.(2020)Han, Zegras, Pereira, and Ben-Akiva]{han2020neural}
Y.~Han, C.~Zegras, F.~C. Pereira, and M.~Ben-Akiva.
\newblock A neural-embedded choice model: Tastenet-mnl modeling taste
  heterogeneity with flexibility and interpretability.
\newblock \emph{arXiv preprint arXiv:2002.00922}, 2020.

\bibitem[He et~al.(2016)He, Zhang, Ren, and Sun]{he2016deep}
K.~He, X.~Zhang, S.~Ren, and J.~Sun.
\newblock Deep residual learning for image recognition.
\newblock In \emph{Proceedings of the IEEE conference on computer vision and
  pattern recognition}, pages 770--778, 2016.

\bibitem[Hensher and Ton(2000)]{hensher2000comparison}
D.~A. Hensher and T.~T. Ton.
\newblock A comparison of the predictive potential of artificial neural
  networks and nested logit models for commuter mode choice.
\newblock \emph{Transportation Research Part E: Logistics and Transportation
  Review}, 36\penalty0 (3):\penalty0 155--172, 2000.

\bibitem[Hruschka(2007)]{hruschka2007using}
H.~Hruschka.
\newblock Using a heterogeneous multinomial probit model with a neural net
  extension to model brand choice.
\newblock \emph{Journal of Forecasting}, 26\penalty0 (2):\penalty0 113--127,
  2007.

\bibitem[Hruschka et~al.(2002)Hruschka, Fettes, Probst, and
  Mies]{hruschka2002flexible}
H.~Hruschka, W.~Fettes, M.~Probst, and C.~Mies.
\newblock A flexible brand choice model based on neural net methodology a
  comparison to the linear utility multinomial logit model and its latent class
  extension.
\newblock \emph{OR spectrum}, 24\penalty0 (2):\penalty0 127--143, 2002.

\bibitem[Hruschka et~al.(2004)Hruschka, Fettes, and
  Probst]{hruschka2004empirical}
H.~Hruschka, W.~Fettes, and M.~Probst.
\newblock An empirical comparison of the validity of a neural net based
  multinomial logit choice model to alternative model specifications.
\newblock \emph{European Journal of Operational Research}, 159\penalty0
  (1):\penalty0 166--180, 2004.

\bibitem[Iranitalab and Khattak(2017)]{iranitalab2017comparison}
A.~Iranitalab and A.~Khattak.
\newblock Comparison of four statistical and machine learning methods for crash
  severity prediction.
\newblock \emph{Accident Analysis \& Prevention}, 108:\penalty0 27--36, 2017.

\bibitem[Jin et~al.(2020)Jin, Hillel, Elshafie, and
  Bierlaire]{jin2020systematic}
Y.~Jin, T.~Hillel, M.~Elshafie, and M.~Bierlaire.
\newblock A systematic review of machine learning classification methodologies
  for modelling passenger mode choice.
\newblock \emph{Journal of Choice Modelling}, 2020.

\bibitem[Karlaftis and Vlahogianni(2011)]{karlaftis2011statistical}
M.~G. Karlaftis and E.~I. Vlahogianni.
\newblock Statistical methods versus neural networks in transportation
  research: Differences, similarities and some insights.
\newblock \emph{Transportation Research Part C: Emerging Technologies},
  19\penalty0 (3):\penalty0 387--399, 2011.

\bibitem[Kim et~al.(2016)Kim, Rasouli, and Timmermans]{kim2016}
J.~Kim, S.~Rasouli, and H.~Timmermans.
\newblock A hybrid choice model with a nonlinear utility function and bounded
  distribution for latent variables: application to purchase intention
  decisions of electric cars.
\newblock \emph{Transportmetrica A: Transport Science}, 12\penalty0
  (10):\penalty0 909--932, 2016.

\bibitem[Kingma and Ba(2014)]{kingma2014adam}
D.~P. Kingma and J.~Ba.
\newblock Adam: A method for stochastic optimization.
\newblock \emph{arXiv preprint arXiv:1412.6980}, 2014.

\bibitem[Kneib et~al.(2007)Kneib, Baumgartner, and Steiner]{kneib2007}
T.~Kneib, B.~Baumgartner, and W.~J. Steiner.
\newblock Semiparametric multinomial logit models for analysing consumer choice
  behaviour.
\newblock \emph{AStA Advances in Statistical Analysis}, 91\penalty0
  (3):\penalty0 225--244, 2007.

\bibitem[Krogh and Hertz(1992)]{krogh1992simple}
A.~Krogh and J.~A. Hertz.
\newblock A simple weight decay can improve generalization.
\newblock In \emph{Advances in neural information processing systems}, pages
  950--957, 1992.

\bibitem[Lee et~al.(2018)Lee, Derrible, and Pereira]{lee2018comparison}
D.~Lee, S.~Derrible, and F.~C. Pereira.
\newblock Comparison of four types of artificial neural network and a
  multinomial logit model for travel mode choice modeling.
\newblock \emph{Transportation Research Record}, page 0361198118796971, 2018.

\bibitem[Lee(1982)]{lee1982specification}
L.-F. Lee.
\newblock Specification error in multinomial logit models: Analysis of the
  omitted variable bias.
\newblock \emph{Journal of Econometrics}, 20\penalty0 (2):\penalty0 197--209,
  1982.

\bibitem[Lh{\'e}ritier et~al.(2018)Lh{\'e}ritier, Bocamazo, Delahaye, and
  Acuna-Agost]{lheritier2018}
A.~Lh{\'e}ritier, M.~Bocamazo, T.~Delahaye, and R.~Acuna-Agost.
\newblock Airline itinerary choice modeling using machine learning.
\newblock \emph{Journal of Choice Modelling}, 2018.

\bibitem[Li et~al.(2019)Li, Lin, and Shen]{li2019deep}
Q.~Li, T.~Lin, and Z.~Shen.
\newblock Deep learning via dynamical systems: An approximation perspective.
\newblock \emph{arXiv preprint arXiv:1912.10382}, 2019.

\bibitem[Liao et~al.(2018)Liao, Zhang, Cai, Tang, Gao, Wu, Yang, Zhu, Guo, and
  Wu]{liao2018dest}
B.~Liao, J.~Zhang, M.~Cai, S.~Tang, Y.~Gao, C.~Wu, S.~Yang, W.~Zhu, Y.~Guo, and
  F.~Wu.
\newblock Dest-resnet: A deep spatiotemporal residual network for hotspot
  traffic speed prediction.
\newblock In \emph{Proceedings of the 26th ACM international conference on
  Multimedia}, pages 1883--1891, 2018.

\bibitem[Likert(1932)]{likert1932technique}
R.~Likert.
\newblock A technique for the measurement of attitudes.
\newblock \emph{Archives of psychology}, 1932.

\bibitem[McFadden(1974)]{mcfadden1974}
D.~McFadden.
\newblock The measurement of urban travel demand.
\newblock \emph{Journal of public economics}, 3\penalty0 (4):\penalty0
  303--328, 1974.

\bibitem[McFadden(1978)]{mcfadden1978modeling}
D.~McFadden.
\newblock Modeling the choice of residential location.
\newblock \emph{Transportation Research Record}, \penalty0 (673), 1978.

\bibitem[Mohammadian and Miller(2002)]{mohammadian2002nested}
A.~Mohammadian and E.~Miller.
\newblock Nested logit models and artificial neural networks for predicting
  household automobile choices: comparison of performance.
\newblock \emph{Transportation Research Record: Journal of the Transportation
  Research Board}, \penalty0 (1807):\penalty0 92--100, 2002.

\bibitem[Murdoch et~al.(2019)Murdoch, Singh, Kumbier, Abbasi-Asl, and
  Yu]{murdoch2019definitions}
W.~J. Murdoch, C.~Singh, K.~Kumbier, R.~Abbasi-Asl, and B.~Yu.
\newblock Definitions, methods, and applications in interpretable machine
  learning.
\newblock \emph{Proceedings of the National Academy of Sciences}, 116\penalty0
  (44):\penalty0 22071--22080, 2019.

\bibitem[Nam et~al.(2017)Nam, Kim, Cho, and Jayakrishnan]{nam2017}
D.~Nam, H.~Kim, J.~Cho, and R.~Jayakrishnan.
\newblock A model based on deep learning for predicting travel mode choice.
\newblock In \emph{Proceedings of the Transportation Research Board 96th Annual
  Meeting Transportation Research Board, Washington, DC, USA}, pages 8--12,
  2017.

\bibitem[Omrani(2015)]{omrani2015}
H.~Omrani.
\newblock Predicting travel mode of individuals by machine learning.
\newblock \emph{Transportation Research Procedia}, 10:\penalty0 840--849, 2015.

\bibitem[Otsuka and Osogami(2016)]{otsuka2016deep}
M.~Otsuka and T.~Osogami.
\newblock A deep choice model.
\newblock In \emph{AAAI}, pages 850--856, 2016.

\bibitem[Paredes et~al.(2017)Paredes, Hemberg, O'Reilly, and
  Zegras]{paredes2017machine}
M.~Paredes, E.~Hemberg, U.-M. O'Reilly, and C.~Zegras.
\newblock Machine learning or discrete choice models for car ownership demand
  estimation and prediction?
\newblock In \emph{Models and Technologies for Intelligent Transportation
  Systems (MT-ITS), 2017 5th IEEE International Conference on}, pages 780--785.
  IEEE, 2017.

\bibitem[Pekel and Soner~Kara(2017)]{pekel2017}
E.~Pekel and S.~Soner~Kara.
\newblock A comprehensive review for artificial neural network application to
  public transportation.
\newblock \emph{Sigma: Journal of Engineering \& Natural
  Sciences/M{\"u}hendislik ve Fen Bilimleri Dergisi}, 35\penalty0 (1), 2017.

\bibitem[Pirra and Diana(2018)]{pirra2018}
M.~Pirra and M.~Diana.
\newblock A study of tour-based mode choice based on a support vector machine
  classifier.
\newblock \emph{Transportation Planning and Technology}, pages 1--14, 2018.

\bibitem[Ribeiro et~al.(2016)Ribeiro, Singh, and Guestrin]{ribeiro2016should}
M.~T. Ribeiro, S.~Singh, and C.~Guestrin.
\newblock " why should i trust you?" explaining the predictions of any
  classifier.
\newblock In \emph{Proceedings of the 22nd ACM SIGKDD international conference
  on knowledge discovery and data mining}, pages 1135--1144, 2016.

\bibitem[Rish et~al.(2001)]{rish2001}
I.~Rish et~al.
\newblock An empirical study of the naive bayes classifier.
\newblock In \emph{IJCAI 2001 workshop on empirical methods in artificial
  intelligence}, volume~3, pages 41--46. IBM New York, 2001.

\bibitem[Samek et~al.(2016)Samek, Binder, Montavon, Lapuschkin, and
  M{\"u}ller]{samek2016evaluating}
W.~Samek, A.~Binder, G.~Montavon, S.~Lapuschkin, and K.-R. M{\"u}ller.
\newblock Evaluating the visualization of what a deep neural network has
  learned.
\newblock \emph{IEEE transactions on neural networks and learning systems},
  28\penalty0 (11):\penalty0 2660--2673, 2016.

\bibitem[Sayed and Razavi(2000)]{sayed2000comparison}
T.~Sayed and A.~Razavi.
\newblock Comparison of neural and conventional approaches to mode choice
  analysis.
\newblock \emph{Journal of Computing in Civil Engineering}, 14\penalty0
  (1):\penalty0 23--30, 2000.

\bibitem[Schindler et~al.(2007)Schindler, Baumgartner, and
  Hruschka]{schindler2007}
M.~Schindler, B.~Baumgartner, and H.~Hruschka.
\newblock Nonlinear effects in brand choice models: comparing heterogeneous
  latent class to homogeneous nonlinear models.
\newblock \emph{Schmalenbach Business Review}, 59\penalty0 (2):\penalty0
  118--137, 2007.

\bibitem[Shannon(1948)]{shannon1948mathematical}
C.~E. Shannon.
\newblock A mathematical theory of communication.
\newblock \emph{Bell system technical journal}, 27\penalty0 (3):\penalty0
  379--423, 1948.

\bibitem[Shen(2009)]{shen2009}
J.~Shen.
\newblock Latent class model or mixed logit model? {A} comparison by transport
  mode choice data.
\newblock \emph{Applied Economics}, 41\penalty0 (22):\penalty0 2915--2924,
  2009.

\bibitem[Shmueli et~al.(1996)Shmueli, Salomon, and Shefer]{shmueli1996}
D.~Shmueli, I.~Salomon, and D.~Shefer.
\newblock Neural network analysis of travel behavior: evaluating tools for
  prediction.
\newblock \emph{Transportation Research Part C: Emerging Technologies},
  4\penalty0 (3):\penalty0 151--166, 1996.

\bibitem[Shrikumar et~al.(2017)Shrikumar, Greenside, and
  Kundaje]{shrikumar2017learning}
A.~Shrikumar, P.~Greenside, and A.~Kundaje.
\newblock Learning important features through propagating activation
  differences, 2017.

\bibitem[Simonyan et~al.(2013)Simonyan, Vedaldi, and
  Zisserman]{simonyan2013deep}
K.~Simonyan, A.~Vedaldi, and A.~Zisserman.
\newblock Deep inside convolutional networks: Visualising image classification
  models and saliency maps.
\newblock \emph{arXiv preprint arXiv:1312.6034}, 2013.

\bibitem[Torres et~al.(2011)Torres, Hanley, and Riera]{torres2011}
C.~Torres, N.~Hanley, and A.~Riera.
\newblock How wrong can you be? {I}mplications of incorrect utility function
  specification for welfare measurement in choice experiments.
\newblock \emph{Journal of Environmental Economics and Management}, 62\penalty0
  (1):\penalty0 111--121, 2011.

\bibitem[van Cranenburgh and Alwosheel(2019)]{Sander2019}
S.~van Cranenburgh and A.~Alwosheel.
\newblock An artificial neural network based approach to investigate
  travellers' decision rules.
\newblock \emph{Transportation Research Part C: Emerging Technologies},
  98:\penalty0 152--166, 2019.

\bibitem[Van Der~Pol et~al.(2014)Van Der~Pol, Currie, Kromm, and Ryan]{van2014}
M.~Van Der~Pol, G.~Currie, S.~Kromm, and M.~Ryan.
\newblock Specification of the utility function in discrete choice experiments.
\newblock \emph{Value in Health}, 17\penalty0 (2):\penalty0 297--301, 2014.

\bibitem[Vij et~al.(2013)Vij, Carrel, and Walker]{vij2013}
A.~Vij, A.~Carrel, and J.~L. Walker.
\newblock Incorporating the influence of latent modal preferences on travel
  mode choice behavior.
\newblock \emph{Transportation Research Part A: Policy and Practice},
  54:\penalty0 164--178, 2013.

\bibitem[von Rueden et~al.(2020)von Rueden, Mayer, Beckh, Georgiev,
  Giesselbach, Heese, Kirsch, Pfrommer, Pick, Ramamurthy,
  et~al.]{von2020informed}
L.~von Rueden, S.~Mayer, K.~Beckh, B.~Georgiev, S.~Giesselbach, R.~Heese,
  B.~Kirsch, J.~Pfrommer, A.~Pick, R.~Ramamurthy, et~al.
\newblock Informed machine learning--a taxonomy and survey of integrating
  knowledge into learning systems.
\newblock \emph{arXiv preprint arXiv:1903.12394}, 2020.

\bibitem[Vovsha(1997)]{vovsha1997application}
P.~Vovsha.
\newblock Application of cross-nested logit model to mode choice in tel aviv,
  israel, metropolitan area.
\newblock \emph{Transportation Research Record}, 1607\penalty0 (1):\penalty0
  6--15, 1997.

\bibitem[Wang et~al.(2018{\natexlab{a}})Wang, Yuan, Shi, and
  Osher]{wang2018enresnet}
B.~Wang, B.~Yuan, Z.~Shi, and S.~J. Osher.
\newblock Enresnet: Resnet ensemble via the feynman-kac formalism.
\newblock \emph{arXiv preprint arXiv:1811.10745}, 2018{\natexlab{a}}.

\bibitem[Wang et~al.(2018{\natexlab{b}})Wang, Wang, Bailey, and
  Zhao]{wang2018framing}
S.~Wang, Q.~Wang, N.~Bailey, and J.~Zhao.
\newblock Deep neural networks for choice analysis: A statistical learning
  theory perspective, 2018{\natexlab{b}}.

\bibitem[Wang et~al.(2018{\natexlab{c}})Wang, Wang, and Zhao]{wang2018using}
S.~Wang, Q.~Wang, and J.~Zhao.
\newblock Deep neural networks for choice analysis: Extracting complete
  economic information for interpretation, 2018{\natexlab{c}}.

\bibitem[Wang et~al.(2020)Wang, Mo, and Zhao]{wang2020deep}
S.~Wang, B.~Mo, and J.~Zhao.
\newblock Deep neural networks for choice analysis: Architecture design with
  alternative-specific utility functions.
\newblock \emph{Transportation Research Part C: Emerging Technologies},
  112:\penalty0 234--251, 2020.

\bibitem[West et~al.(1997)West, Brockett, and Golden]{west1997comparative}
P.~M. West, P.~L. Brockett, and L.~L. Golden.
\newblock A comparative analysis of neural networks and statistical methods for
  predicting consumer choice.
\newblock \emph{Marketing Science}, pages 370--391, 1997.

\bibitem[Williams(1977)]{williams1977formation}
H.~C. Williams.
\newblock On the formation of travel demand models and economic evaluation
  measures of user benefit.
\newblock \emph{Environment and planning A}, 9\penalty0 (3):\penalty0 285--344,
  1977.

\bibitem[Wong and Farooq(2019)]{wong2019reslogit}
M.~Wong and B.~Farooq.
\newblock Reslogit: A residual neural network logit model.
\newblock \emph{arXiv preprint arXiv:1912.10058}, 2019.

\bibitem[Wong and Farooq(2020)]{wong2020bi}
M.~Wong and B.~Farooq.
\newblock A bi-partite generative model framework for analyzing and simulating
  large scale multiple discrete-continuous travel behaviour data.
\newblock \emph{Transportation Research Part C: Emerging Technologies},
  110:\penalty0 247--268, 2020.

\bibitem[Wong et~al.(2018)Wong, Farooq, and Bilodeau]{wong2018}
M.~Wong, B.~Farooq, and G.-A. Bilodeau.
\newblock Discriminative conditional restricted boltzmann machine for discrete
  choice and latent variable modelling.
\newblock \emph{Journal of choice modelling}, 29:\penalty0 152--168, 2018.

\bibitem[Xiong and Mannering(2013)]{xiong2013}
Y.~Xiong and F.~L. Mannering.
\newblock The heterogeneous effects of guardian supervision on adolescent
  driver-injury severities: A finite-mixture random-parameters approach.
\newblock \emph{Transportation research part B: methodological}, 49:\penalty0
  39--54, 2013.

\bibitem[Zhao et~al.(2020)Zhao, Yan, Yu, and
  Van~Hentenryck]{zhao2020prediction}
X.~Zhao, X.~Yan, A.~Yu, and P.~Van~Hentenryck.
\newblock Prediction and behavioral analysis of travel mode choice: A
  comparison of machine learning and logit models.
\newblock \emph{Travel Behaviour and Society}, 20:\penalty0 22--35, 2020.

\bibitem[Zhou et~al.(2019)Zhou, Wang, and Li]{zhou2019bike}
X.~Zhou, M.~Wang, and D.~Li.
\newblock Bike-sharing or taxi? modeling the choices of travel mode in chicago
  using machine learning.
\newblock \emph{Journal of transport geography}, 79:\penalty0 102479, 2019.

\end{thebibliography}

\end{document}